%% file: main.tex
\documentclass[twocolumn,letterpaper,10pt]{article}
\pdfoutput=1
\usepackage[pagenumbers]{iccv} 

\usepackage{url}
\usepackage[utf8]{inputenc} 
\usepackage{booktabs}       
\usepackage{amsfonts}       
\usepackage{nicefrac}       
\usepackage{xcolor}         
\usepackage{multirow}
\usepackage{dsfont}
\usepackage{soul}
\usepackage{adjustbox}
\usepackage{xspace}
\usepackage{bm}
\usepackage{bbm}
\usepackage{amsmath}

\usepackage{anyfontsize}

\usepackage{algpseudocode}                  
\usepackage{setspace}                       
\usepackage{scrextend}
\usepackage{mathtools}
\usepackage{graphicx}
\usepackage{subcaption}
\usepackage{enumitem}
\usepackage{colortbl}
\usepackage{wrapfig}
\usepackage{kotex}
\usepackage{lipsum}
\usepackage{xcolor,pifont}

\usepackage{duckuments}
\usepackage{makecell}
\usepackage[ruled,vlined]{algorithm2e}
\usepackage{flushend}
\usepackage{listings}
\usepackage[frozencache,cachedir=.]{minted}

\usepackage{microtype}
\usepackage[T1]{fontenc}
\usepackage{zi4} 

\usepackage{tikz}
\usetikzlibrary{fit,calc}

\newcommand{\boxit}[2]{
    \tikz[remember picture,overlay] \node (A) {}; \ignorespaces
    \tikz[remember picture,overlay]{
        \node[yshift=3pt, fill=#1, opacity=.25,
        fit={($(A)+(0,0.3\baselineskip)$)($(A)+(0.78\linewidth,-{#2}\baselineskip - 0.15\baselineskip)$)}] {};}\ignorespaces
}

\usepackage{soul} 
\newcommand{\ctext}[2]{%
  \begingroup
  \sethlcolor{#1}%
  \hl{#2}%
  \endgroup
}

\newcommand{\lname}{Synchronized Coupled Sampling\xspace} 
\newcommand{\sname}{SynCoS\xspace} 

\definecolor{cvprblue}{rgb}{0.21,0.49,0.74}
\definecolor{pinegreen}{rgb}{0.0, 0.47, 0.44}
\definecolor{cornellred}{rgb}{0.7, 0.11, 0.11}
\definecolor{cadmiumgreen}{rgb}{0.0, 0.42, 0.24}
\definecolor{spirodiscoball}{rgb}{0.06, 0.75, 0.99}
\definecolor{Red7}{rgb}{0.941, 0.243, 0.243}
\definecolor{Eqpink}{RGB}{241,241,214}
\definecolor{Green7}{RGB}{55, 178, 77}

\newcommand{\cmark}{\ding{51}}%
\newcommand{\xmark}{\ding{55}}
\newcommand{\ck}{\color{Green7}{\cmark}}
\newcommand{\xk}{\color{Red7}{\xmark}}

\usepackage{tikz}

\algrenewcommand\algorithmicrequire{\textbf{Input:}}
\algrenewcommand\algorithmicensure{\textbf{Output:}}

\definecolor{iccvblue}{rgb}{0.21,0.49,0.74}
\usepackage[pagebackref,breaklinks,colorlinks,allcolors=iccvblue]{hyperref}

\title{Tuning-Free Multi-Event Long Video Generation \\ via Synchronized Coupled Sampling
}

\author{
Subin Kim$^{1}$ 
\qquad Seoung Wug Oh$^{2}$ 
\qquad Jui-Hsien Wang$^{2}$ 
\qquad Joon-Young Lee$^{2}$
\qquad Jinwoo Shin$^{1}$\\[0.5em]
$^{1}$KAIST \qquad $^{2}$Adobe Research\\[0.3em]
{\tt\small {subin-kim}@kaist.ac.kr}
}

\lstdefinestyle{mystyle}{
    basicstyle=\ttfamily\footnotesize,  
    backgroundcolor=\color{background},   
    basicstyle=\ttfamily\footnotesize,    
    keywordstyle=\color{keyword}\bfseries, 
    commentstyle=\color{comment}\itshape, 
    stringstyle=\color{string},           
    numberstyle=\tiny\color{gray},        
    numbers=left,                         
    numbersep=5pt,                        
    frame=single,                         
    breaklines=true,                      
    captionpos=b,                         
    showspaces=false,                     
    showstringspaces=false,               
    showtabs=false,                       
    tabsize=4                             
}

\begin{document}

\input{resources/teaser}

\input{sec/0_abstract}    
\input{sec/1_intro}
\input{sec/3_preliminaries}
\input{sec/4_motivation}
\input{sec/5_method}

\input{sec/6_exp}
\input{sec/2_related}
\input{sec/7_conclusion}

\clearpage

{
    \small
    \bibliographystyle{ieeenat_fullname}
    \bibliography{main}
}

\appendix
\input{sec/X_suppl}

\end{document}

%% file: resources/teaser.tex
\twocolumn[{
\renewcommand\twocolumn[1][]{#1}
\maketitle
\vspace{-0.4in}
    \makebox[\textwidth]{\includegraphics[width=0.98\textwidth]{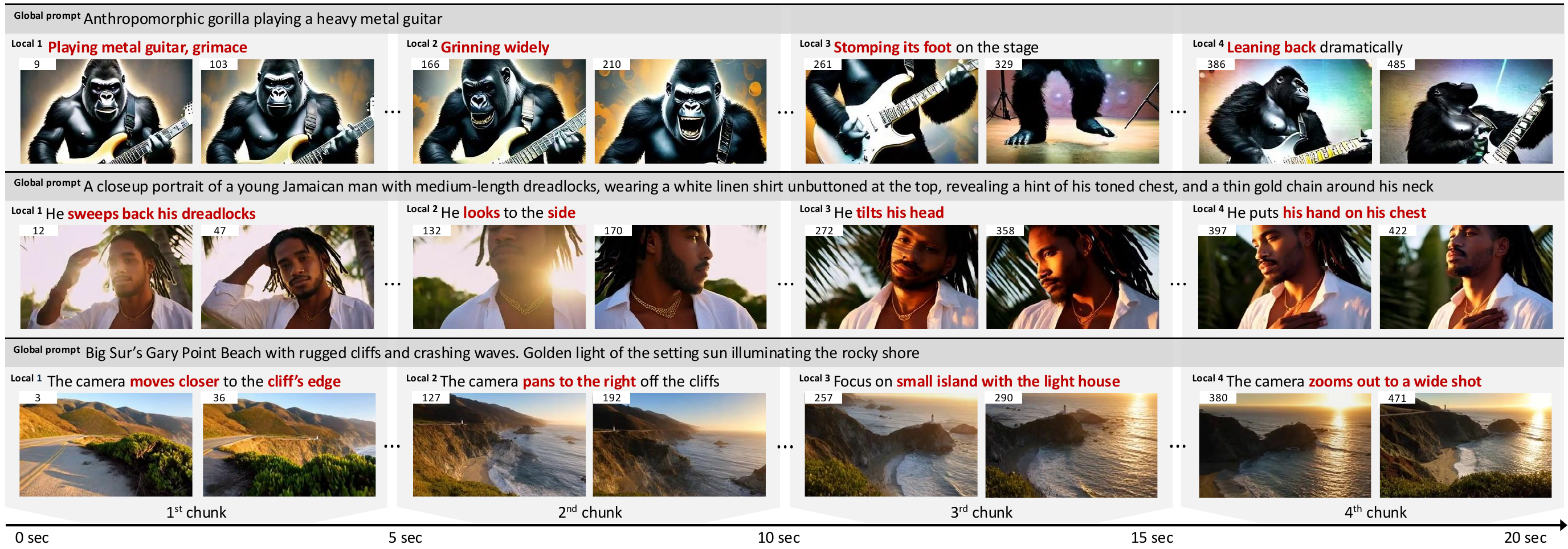}}
    \vspace{-0.3in}
    \captionof{figure}{\textbf{Multi-event long video generation results} using our tuning-free inference framework, \sname. 
    Each example is around 21 seconds of video at 24 fps (4$\times$ longer than the base model ).
    Frame indices are displayed in each frame.
    \sname generates high-quality, long videos with multi-event dynamics while achieving both seamless transitions between frames and long-term semantic consistency throughout. 
    }
    \label{fig:teaser}
}
\vspace{0.1in}
]

%% file: sec/0_abstract.tex
\begin{abstract}
While recent advancements in text-to-video diffusion models enable high-quality short video generation from a single prompt, generating real-world long videos in a single pass remains challenging due to limited data and high computational costs.
To address this, several works propose tuning-free approaches, i.e., extending existing models for long video generation, specifically using multiple prompts to allow for dynamic and controlled content changes. 
However, these methods primarily focus on ensuring smooth transitions between adjacent frames, often leading to content drift and a gradual loss of semantic coherence over longer sequences.
To tackle such an issue, we propose Synchronized Coupled Sampling (SynCoS), a novel inference framework that synchronizes denoising paths across the entire video, ensuring long-range consistency across both adjacent and distant frames.
Our approach combines two complementary sampling strategies: reverse and optimization-based sampling, which ensure seamless local transitions and enforce global coherence, respectively.
However, directly alternating between these samplings misaligns denoising trajectories, disrupting prompt guidance and introducing unintended content changes as they operate independently.
To resolve this, SynCoS synchronizes them through a grounded timestep and a fixed baseline noise, ensuring fully coupled sampling with aligned denoising paths.
Extensive experiments show that SynCoS significantly improves multi-event long video generation, achieving smoother transitions and superior long-range coherence, outperforming previous approaches both quantitatively and qualitatively.
\footnote{Visualizations are available at: \url{https://syncos2025.github.io/}.}
\end{abstract}

%% file: sec/1_intro.tex
\input{resources/emph}

\section{Introduction}\label{sec:intro}

Recently, text-to-video (T2V) diffusion models have shown impressive capabilities in synthesizing high-quality videos from a single text prompt~\citep{polyak2024movie, girdhar2023emu, bar2024lumiere, yang2024cogvideox, lin2024open, jin2024pyramidal}.
However, these models primarily generate short, fixed-length videos that capture a single event, mainly due to high computational training costs and limited high-quality long video data~\citep{wu2024mind, villegas2022phenaki, oh2023mtvg}.
In contrast, real-world videos are long and comprise multiple events that introduce dynamics, such as object motion and camera movement, all within a coherent scenario.
Thus, extending T2V models for long video generation requires handling event transitions while ensuring global coherence throughout the video.

Existing methods often struggle to achieve this balance, failing to enforce long-range consistency.
Training-based approaches introduce conditioning modules~\citep{ge2022long, villegas2022phenaki, tian2024videotetris} to generate video chunks autoregressively.
However, relying solely on previously generated chunks makes maintaining a coherent global structure difficult, and repeated conditioning leads to error accumulation and degraded frame quality.

Alternatively, several works~\citep{oh2023mtvg, kim2024fifo, wang2023gen, qiu2023freenoise, lu2025freelong, cai2024ditctrl} propose inference techniques to extend T2V diffusion models, avoiding additional training and error accumulation.
One common and intuitive approach is fusion, which smoothly connects short videos into a longer one by dividing a long video into overlapping chunks, denoising each with different prompts, and applying local fusion to overlapped regions for seamless transitions.
However, these methods focus only on local smoothness, failing to integrate information across distant chunks.
This issue becomes even more pronounced in multi-prompt scenarios, where denoising paths between chunks diverge more significantly than with a single prompt, resulting in severe inconsistencies in semantics, and style across the generated video (see $1^{\text{st}}$ and $2^{\text{nd}}$ rows in Figure~\ref{fig:emph}).

While local fusion-based inference techniques are promising as they require no additional training, they lack a mechanism to enforce long-range consistency. 
We argue that maintaining a shared denoising trajectory across both short- and long-distance chunks is essential for generating coherent long videos.
To address this, we explore an optimization-based sampling approach: Collaborative Score Distillation (CSD)~\citep{kim2023collaborative}, which synchronizes information across adjacent and distant samples, enforcing long-range consistency.
However, when directly applied to long video generation, CSD completely fails to produce high-quality videos despite aligning denoising trajectories across chunks (see Section~\ref{motivation:fusion_csd}).

In this paper, we propose Synchronized Coupled Sampling (\sname), a novel inference framework for extending any T2V diffusion model to multi-event long video generation.
Our key idea is to jointly preserve local smoothness and global coherence by coupling two complementary samplings. 
However, a direct combination misaligns their denoising trajectories, weakening prompt guidance and causing unintended content variations. 
To address this, we introduce a synchronized mechanism when coupling them.

Specifically, \sname integrates Denoising Diffusion Implicit Models (DDIM)~\citep{song2020denoising}, built on fusion, to enforce smooth local transitions between adjacent chunks while leveraging CSD to maintain long-range coherence across chunks.
To enable synchronous operation when coupling these sampling methods, \sname introduces a grounded timestep and fixed baseline noise, ensuring alignment across the entire denoising process.
Furthermore, \sname proposes a structured prompt to ensure a coherent long scenario with dynamics, combining a global prompt for scenario-wide consistency with local prompts for fine-grained controls.

This newly introduced coupled sampling, along with its key synchronization components—grounded timestep and fixed baseline noise—and the structured prompt, defines \sname as a new inference framework.
As a unified inference framework, \sname ensures that local smoothness and global coherence work in tandem, enabling coherent long video generation with multi-event dynamics.

We evaluate \sname across various challenging scenarios in long video generation, including multiple events (e.g., object motion, compositions, camera movements, background changes, etc.) across different denoising models to comprehensively verify its efficacy.
Consistently, \sname significantly outperforms existing tuning-free methods in temporal consistency, video quality, and prompt fidelity.

\vspace{0.02in}
\noindent\textbf{Contributions}. Our main contributions are as follows:
\begin{itemize}[topsep=0.0pt,itemsep=1.2pt,leftmargin=2.5mm]

\item We propose \sname, a novel inference framework that extends any T2V diffusion model for multi-event long video generation without additional tuning.

\item \sname synchronizes two complementary sampling methods to ensure both local transitions and global coherence, introducing a grounded timestep and fixed baseline noise to couple them into a new, unified sampling.

\item \sname incorporates 
a structured prompt for dynamic yet semantically consistent multi-event generation.

\item We extensively validate \sname on various T2V models across diverse long video generation scenarios, achieving state-of-the-art temporal consistency and prompt fidelity.

\end{itemize}

%% file: resources/emph.tex
\begin{figure*}[ht]
\centering\small
\vspace{-0.3in}
\includegraphics[width=0.94\textwidth]{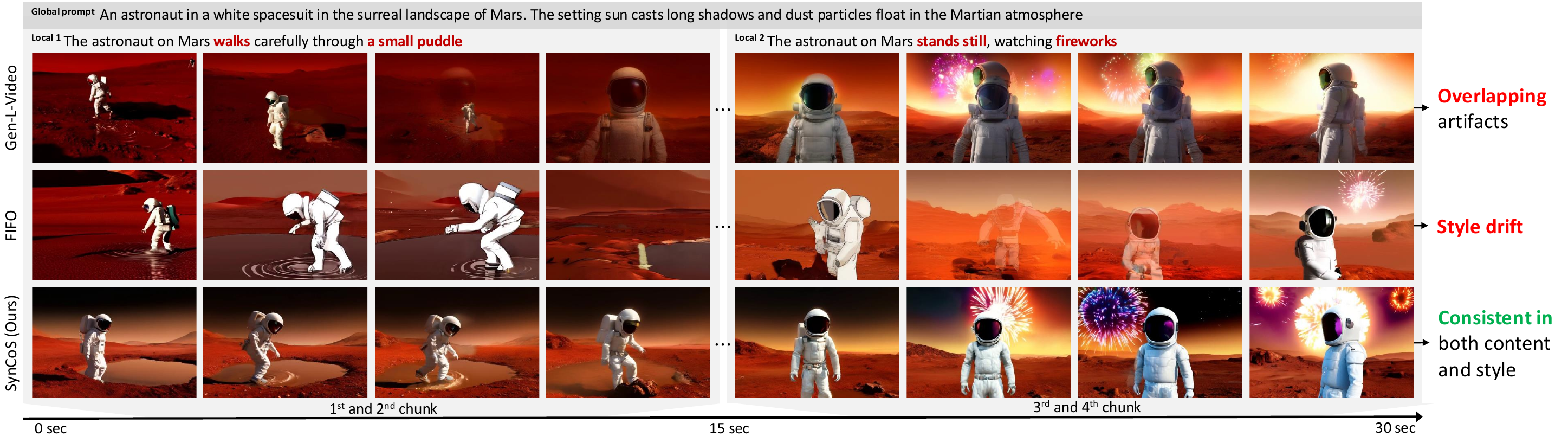}
    \vspace{-0.14in}
\caption{\textbf{Qualitative comparisons} on CogVideoX-2B~\citep{yang2024cogvideox}. 
All examples are 5 times longer in duration compared to the underlying base model, generating a 30-second video.
Unlike Gen-L-Video~\citep{wang2023gen} and FIFO-Diffusion~\citep{kim2024fifo}, which often struggle with overlapping artifacts and style drift, our method, \sname, ensures consistency in both content and style throughout the entire video.
Additionally, \sname generates long videos where each frame faithfully follows its designated prompt while ensuring seamless transitions between frames.
}\label{fig:emph}.
\vspace{-0.3in}
\end{figure*}

%% file: sec/3_preliminaries.tex
\section{Background}\label{sec:pre}
In this section, we present the preliminaries (Section~\ref{sec:prelim}) and basic concept of fusion for extending T2V diffusion models in a tuning-free manner (Section~\ref{motivation:fusion}).

\subsection{Preliminaries}\label{sec:prelim}
We provide a brief overview of diffusion models, score distillation samplings~\citep{poole2022dreamfusion, kim2023collaborative} as the foundation of our method.
For a detailed explanation, refer to Appendix~\ref{appen:derivations}.

\vspace{0.05in}
\noindent\textbf{Diffusion models.} 
Diffusion models are generative models that learn to gradually denoise random Gaussian noise into structured data by reversing a noise-adding process. 
This generative process can be formulated using stochastic differential equations (SDEs)~\citep{song2020score} or their deterministic counterpart, ordinary differential equations (ODEs)~\citep{lipman2022flow}.

A notable variant of diffusion models is Denoising Diffusion Implicit Models (DDIM) \citep{song2020denoising}, which adopts a non-Markovian approach to accelerate sampling while preserving the marginal distribution.
At each step, DDIM estimates the clean data sample, $\hat{\mathbf{x}}_{0|t}$, following Tweedie's formula from a noisy sample, $\mathbf{x}_{t}$, using a learned noise predictor, $\Phi$.
Then, the posterior distribution, $p(\mathbf{x}_{t-1}|\mathbf{x}_t, \mathbf{x}_0)$, is computed as:

\vspace{-0.1in}
\begin{equation}\label{eqn:ddim}
    \mathbf{x}_{t-1} = \sqrt{\bar\alpha_{t-1}} \hat{\mathbf{x}}_{0|t} + \sqrt{1-\bar\alpha_{t-1}}\tilde{\bm{\epsilon}},
\end{equation}

where $\bar\alpha_{t-1}$ is a pre-defined coefficient that regulates noise schedule over time, and $\tilde{\bm{\epsilon}}$ is a weighted combination of the predicted noise estimate term $\bm{\epsilon}_\Phi$, and a stochastic noise term $\bm{\epsilon} \sim \mathcal{N}(\mathbf{0},\mathbf{I})$, controlled by the hyperparameter $\eta$.

For prompt-conditioned generation, classifier-free guidance (CFG) \citep{ho2022classifier} is widely used. The estimated noise is adjusted as: $\bm{\epsilon}_\Phi^\gamma(\mathbf{x}_t; \mathbf{c}, t) = \bm{\epsilon}_\Phi(\mathbf{x}_t; \varnothing, t) + \gamma [\bm{\epsilon}_\Phi(\mathbf{x}_t; \mathbf{c}, t) - \bm{\epsilon}_\Phi(\mathbf{x}_t; \varnothing, t)]$
, where $\gamma$ controls the guidance strength, $\varnothing$ is the null-text embedding and $\mathbf{c}$ is the conditioning text embedding.

\vspace{0.05in}
\noindent\textbf{Score distillation sampling.} Score Distillation Sampling (SDS)~\citep{poole2022dreamfusion, wang2023score} introduces a novel approach for optimizing differentiable parametric functions using diffusion models as a critic.
By doing so, SDS extends the applicability of text-to-image diffusion models to generate and manipulate more complex visual data, including 3D objects and scenes~\citep{tang2023make, chen2023fantasia3d, melas2023realfusion, tsalicoglou2024textmesh, lin2023magic3d, hertz2023delta}.
To achieve this, SDS formulates generative sampling as an optimization problem, allowing control over the generated output by optimizing the parameters, $\theta$, of a function, $g(\theta)$, to guide generation.
Specifically, given a generated sample, $\mathbf{x} = g(\theta)$, the gradient of the diffusion loss function 
with respect to $\theta$ is computed as:

\vspace{-0.15in}
{\small
\begin{equation}
\begin{aligned} 
\nabla_{\theta} \mathcal{L}_{\tt{SDS}} 
 (\Phi; \mathbf{x} = g(\theta) )  \stackrel{\Delta}{=} \mathbb{E}_{t, \bm{\epsilon}} 
\left[ w(t)\left(\bm{\epsilon}_\Phi^\gamma(\mathbf{x}_t; \mathbf{c}, t) - \bm{\epsilon}
\right) 
\frac{\partial \mathbf{x}}{\partial \theta } \right]. 
 \end{aligned}\label{eqn:sds}
\end{equation}
}

\input{resources/motivation}
\noindent\textbf{Collaborative score distillation sampling.}
SDS (Equation~\ref{eqn:sds}) optimizes a single sample, $\mathbf{x}$.
In contrast, Collaborative Score Distillation (CSD)~\citep{kim2023collaborative} extends SDS by incorporating interactions between multiple samples, $\{\mathbf{x}^{(i)}\}_{i=1}^{N}$, allowing them to influence each other, thereby ensuring inter-sample consistency during optimization.

At each iteration, CSD selects a random timestep, $t$, and samples Gaussian noise for each sample, $\bm{\epsilon}^{(i)} \sim \mathcal{N}(\mathbf{0},\mathbf{I})$.
Then, each sample, $\mathbf{x}^{(i)} = g(\theta_i)$, is optimized using the following objective: 

\vspace{-0.15in}
{\small
\begin{equation}
\begin{aligned}\label{eqn:csd}
    & \nabla_{\theta_i} \mathcal{L}_{\tt{CSD}}\big(\Phi; \mathbf{x}^{(i)} = g(\theta_i) \big) \\
    & \stackrel{\Delta}{=}  \frac{w(t)}{N}\sum_{j=1}^N \Bigg[  k(\mathbf{x}_t^{(j)}, \mathbf{x}_t^{(i)})\Big(\boldsymbol{\epsilon}_\Phi(\mathbf{x}_t^{(i)};\mathbf{c},t) - \boldsymbol{\epsilon}^{(i)}\Big) \\
    & \quad \quad \quad \quad \quad \quad + \nabla_{\mathbf{x}_t^{(j)}} k(\mathbf{x}_t^{(j)}, \mathbf{x}_t^{(i)}) \Bigg] \frac{\partial \mathbf{x}^{(i)}}{\partial \theta_i},
\end{aligned}
\vspace{-0.05in}
\end{equation}
}
where $k$ is a positive definite kernel that measures similarity between samples.
By leveraging inter-sample relationships, CSD adjusts gradient updates based on sample affinity, ensuring that each parameter update affects and is affected by other samples to promote global coherency.

\subsection{Fusion-based tuning-free long video generation}\label{motivation:fusion}

T2V diffusion models generate short, single-event videos in one pass due to high computational costs, making long, dynamic video generation challenging.
Fusion offers a practical solution by leveraging the model’s short-clip generation capability for extended multi-event video synthesis: dividing a long video into overlapping short chunks, processing each chunk with different prompts, and fusing overlapped regions for smooth transitions.

\vspace{0.05in}
\noindent\textbf{Problem formulation.}  
Given a pre-trained T2V model, $\Phi$, that generates a short video sample $\mathbf{x} \in \mathcal{R}^{f \times H \times W \times D}$, conditioned on a single text prompt embedding $\mathbf{c}$, our goal is to extend it to generate a long video $\mathbf{x}' \in \mathcal{R}^{F \times H \times W \times D}$, where $F \gg f$.
A common fusion strategy for a long video generation involves partitioning the long video, $\mathbf{x}'$, into overlapping short video chunks ${\{\mathbf{x}^{(i)}\}}_{i=1}^N \in \mathcal{R}^{f \times H \times W \times D}$, with a temporal stride $s$.
Each chunk is denoised independently, conditioned on evolving text prompts ${\{\mathbf{c}^{(i)}\}}_{i=1}^N$, to generate multi-event dynamics.
These short chunks are then merged into a long video by averaging overlapping regions, normalized by the number of contributing chunks.

%% file: resources/motivation.tex
\begin{figure*}[ht]
\centering\small
\begin{minipage}{0.38\textwidth}
    \centering
    \includegraphics[width=0.85\textwidth]{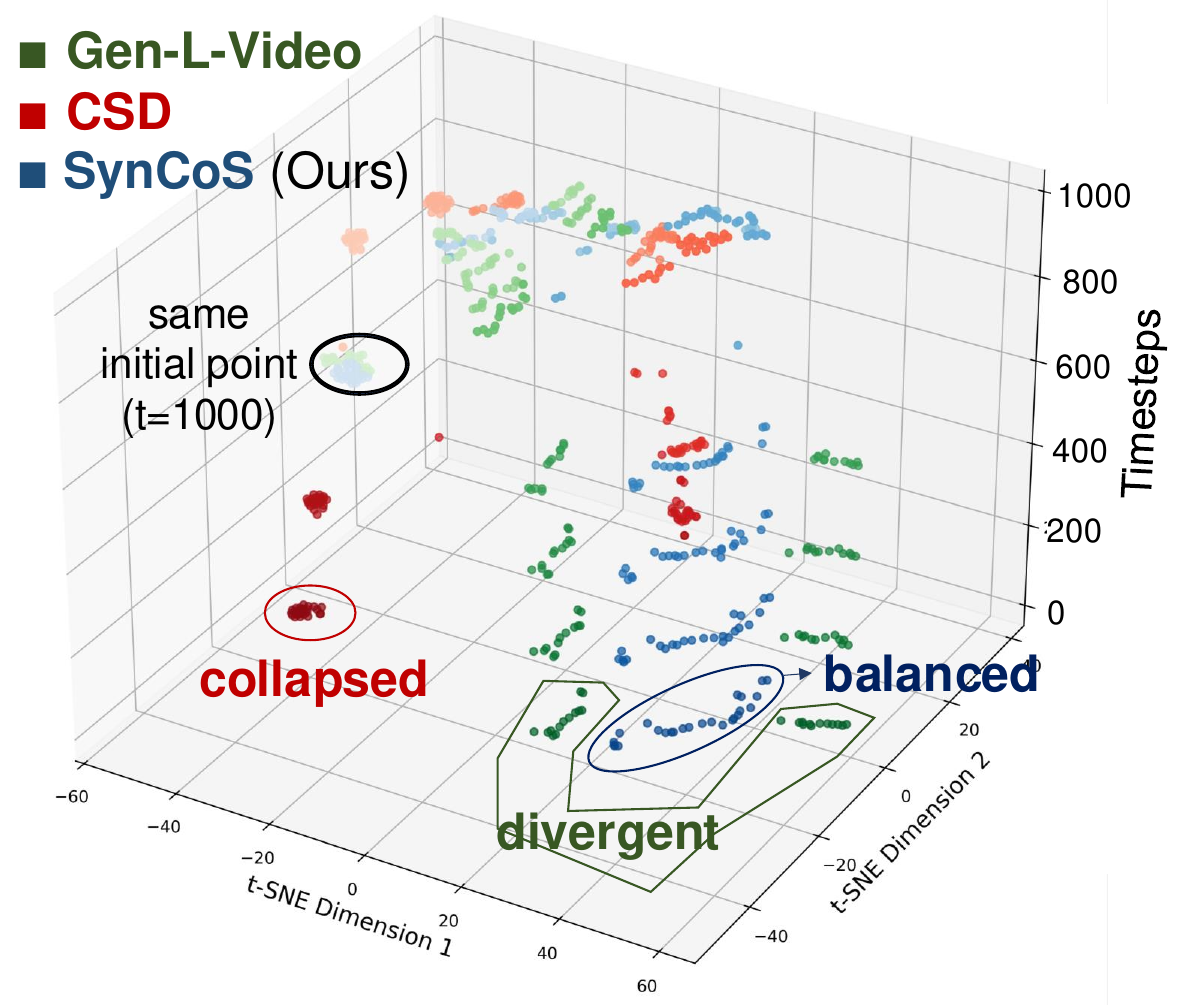}
    \vspace{-0.16in}
    \caption{\textbf{t-SNE visualization} of CLIP~\citep{radford2021learning} features for the predicted video frames $\hat{\mathbf{x}}_{0|t}$, at each timestep using different samplings.
    Faded colors indicate earlier timesteps ($t \approx 1000$), while vivid colors indicate later, small timesteps ($t \approx 0$), illustrating feature trajectory evolution over time (top to bottom).}\label{fig:motiv_tsne}

\end{minipage}
\nolinebreak\hspace{10em minus 10em}
\begin{minipage}{0.60\textwidth}
    \centering
    \vspace{0.2in}
    \includegraphics[width=\textwidth]{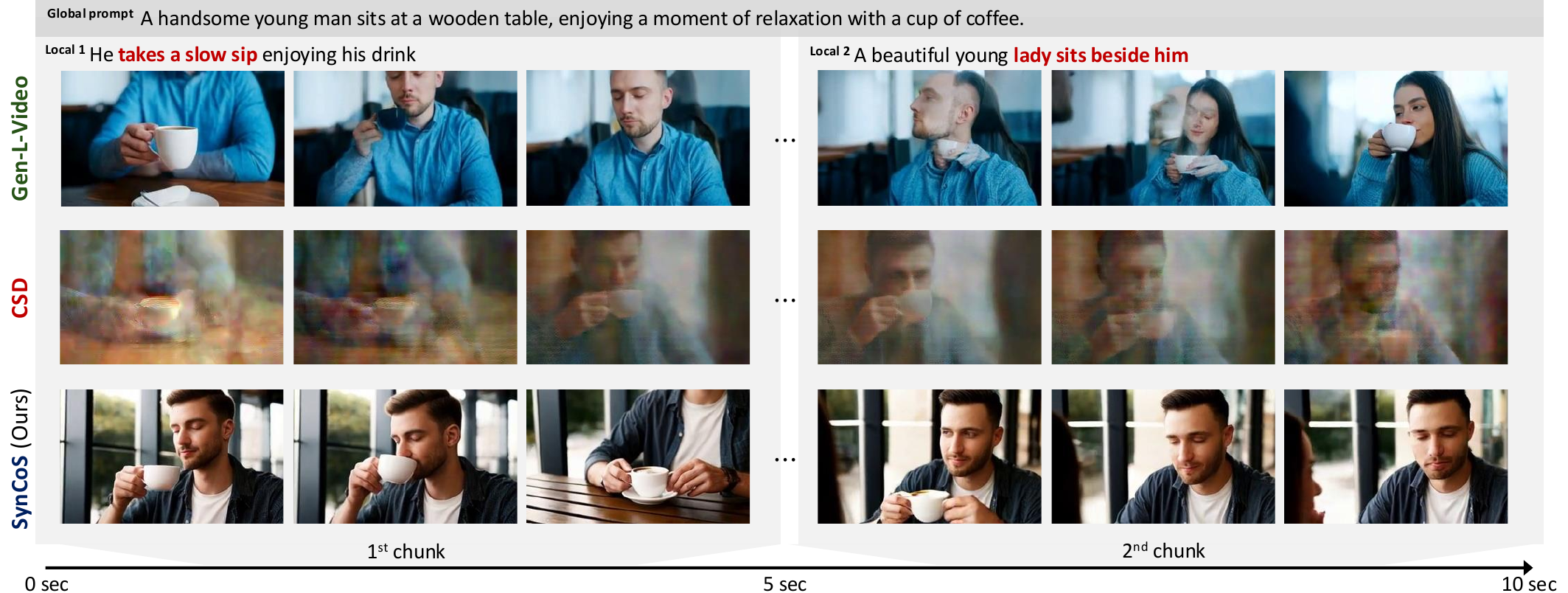}
    \vspace{-0.2in}\caption{\textbf{Qualitative comparison of sampling methods motivating \sname}. Gen-L-Video~\citep{wang2023gen} fails to maintain global coherence, resulting in abrupt appearance changes (\eg, a man morphing into a woman).
    CSD~\citep{kim2023collaborative} retains a similar appearance of a man but shows poor adherence to local prompts, suffering from low frame quality with severe noise-like artifacts.
    In contrast, our method achieves a balance, ensuring high-quality generation, strong prompt fidelity, and temporal coherence.}\label{fig:motiv_qual}
\end{minipage}

\vspace{-0.1in}
\end{figure*}

%% file: sec/4_motivation.tex
\section{Key observations}\label{sec:motivation}

Through fusion, adjacent chunks undergo temporal co-denoising, ensuring smooth transitions.
Gen-L-Video~\citep{wang2023gen}, CSD~\citep{kim2023collaborative}, and our proposed approach all leverage temporal co-denoising via fusion but differ in \emph{what} is fused.
The following sections detail these differences.

\subsection{Local temporal co-denoising with Gen-L-Video}\label{motivation:fusion_genlvideo}

\noindent\textbf{Gen-L-Video fuses posterior distributions.}  
Gen-L-Video~\citep{wang2023gen} extends T2V diffusion models by fusing the posterior distributions of overlapping chunks to maintain local smoothness.
Each chunk undergoes independent denoising using the DDIM sampler, producing intermediate outputs $\mathbf{x}_{t-1}^{(i)}$. 
The final $\mathbf{x}_{t-1}'$ is obtained by fusing the posteriors of overlapping chunks $\mathbf{x}_{t-1}^{(i)}$, as computed in Equation~\ref{eqn:ddim}.

\vspace{0.05in}
\noindent\textbf{Divergence in denoising paths with Gen-L-Video.}
While this local fusion with Gen-L-Video helps maintain smooth transitions between adjacent chunks, it neglects information across distant video chunks to enforce long-range temporal consistency.
As a result, Gen-L-Video produces unnatural transitions, such as abrupt appearance changes (\eg, a man morphing into a woman, as shown in Figure~\ref{fig:motiv_qual}).
We verify this in Figure~\ref{fig:motiv_tsne}, which visualizes how denoising paths evolve over time.
Here, dot color intensity transitions from faded to vivid, representing progressively later timesteps, with dots arranged top to bottom, corresponding to timesteps from $t=1000$ to $t=0$.
The green dots, representing Gen-L-Video, gradually separate, indicating increasing divergence in its denoising paths.

\input{resources/framework}
\subsection{Global temporal co-denoising with CSD}\label{motivation:fusion_csd}
To establish connections between distant video chunks while preserving local smoothness, we explore an optimization-based approach using Collaborative Score Distillation (CSD) for long video generation.
Although CSD was originally designed for visual editing, we apply it to improve global temporal consistency in fusion-based long video generation.

\vspace{0.05in}
\noindent\textbf{Apply CSD for long video generation by fusing its loss.}  
Similar to Gen-L-Video, a long video is divided into overlapping chunks.
The CSD loss for each chunk is computed to synchronize denoising paths across both short- and long-range chunks. Then, the losses in overlapping regions are fused, further reinforcing smooth transitions.

\vspace{0.05in}
\noindent\textbf{Failure of CSD for long video generation.}  
Despite its effectiveness in visual editing, CSD fails when applied to long video generation.
Unlike editing, where an existing source structure guides modifications, video synthesis starts from pure Gaussian noise with no inherent priors.
This lack of structured guidance causes frames to collapse into similar states as denoising progresses (see red dots in Figure~\ref{fig:motiv_tsne}).
As a result, while the generated video retains a similar appearance throughout, it fails to adhere to local prompts and suffers from degraded per-frame quality (see Figure~\ref{fig:motiv_qual}).

%% file: resources/framework.tex
\begin{figure*}[t]
\centering\small
\vspace{-0.2in}
\includegraphics[width=0.98\textwidth]{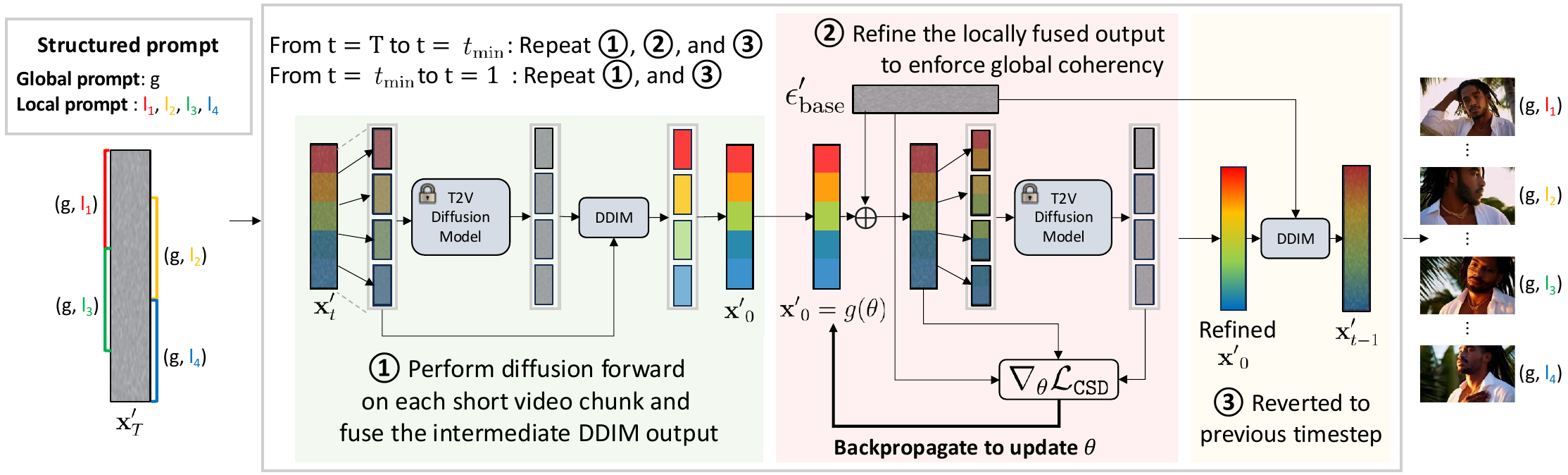}
\vspace{-0.1in}
\caption{
\textbf{Overall illustration of our proposed method, \lname (\sname)}, a tuning-free inference framework for multi-event long video generation. 
\sname performs one-step denoising in three iterative stages, repeated from $t=1000$ to $t=0$.
In the first stage, \sname performs temporal co-denoising with DDIM, dividing the long video into overlapping short chunks, denoising each chunk, and applying fusion for local smoothness.
In the second stage, \sname refines the locally fused output, enforcing global coherence by synchronizing information across both short- and long-distance chunks.
Finally, in the third stage, it reverts the locally and globally refined output to the previous timestep.
Through these three synchronized stages of local and global denoising, \sname ensures smooth transitions, global semantic coherence, and high prompt fidelity, enabling multi-event long video generation.
}\label{fig:framework}
\vspace{-0.2in}
\end{figure*}

%% file: sec/5_method.tex
\section{Method}\label{sec:method}

We propose \lname (\sname), a novel inference framework applicable to any T2V diffusion model for generating temporally consistent, multi-event long videos.
Section~\ref{method:overview} provides an overview of the stages in our coupled sampling, followed by key mechanisms that enable the synchronized coupling of the three stages in Section~\ref{method:component}.

\subsection{\sname: \lname}\label{method:overview}
As discussed in Section~\ref{sec:motivation}, fusion-based temporal co-denoising shows promise for long video generation, but existing methods suffer from different limitations.
Temporal co-denoising with DDIM (\ie, Gen-L-Video) ensures local smoothness but results in divergent denoising paths, whereas temporal co-denoising with CSD facilitates global coherence but results in overly converged, collapsed denoising paths.

To capture both local smoothness and global consistency, \sname combines fusion-based temporal co-denoising with DDIM and CSD at each denoising step.
The key insight is treating the intermediate DDIM output of $\hat{\mathbf{x}}_{0|t}$ as a refinement source for CSD-based optimization, enabling global refinement to build upon locally smoothed updates.
This formulation reframes CSD-based optimization as a progressive refinement process rather than pure generation, where each DDIM update at every denoising step serves as an improved starting point for enforcing global coherence.

\sname consists of three stages, which are repeated at each denoising step (illustrated in Figure~\ref{fig:framework}):

\begin{itemize}[topsep=1.2pt,itemsep=1.2pt,leftmargin=2.5mm]
\vspace{-0.05in}
\item \ctext{green!7}{\textbf{1. Perform temporal co-denoising with DDIM.}} 
Instead of fully reverting each chunk to its previous timestep, $t-1$, \sname computes ${\hat{\mathbf{x}}_{0|t}}^{(i)}$ for every video chunk, then applies fusion to them to produce $\mathbf{x}'_{0}$.

\item \ctext{pink!20}{\textbf{2. Refine the locally fused output to enforce global coherency.}} 
The fused output from the first stage serves as a refinement source, where temporal co-denoising with CSD-based optimization is performed by computing $\nabla_{\mathbf{x}_0^{(i)}} \mathcal{L}_{\tt{CSD}}$ for each chunk and applying fusion on the gradients to update $\mathbf{x}'_{0}$.
This stage synchronizes denoising levels across chunks, effectively managing multi-event scenarios by ensuring coherence when varying prompts cause divergent denoising paths.
By doing so, \sname mitigates content and style drift while preserving global semantic consistency.
Crucially, \sname introduces key components to bridge the first and second stages and the second and final stages, as detailed in Section~\ref{method:component}.

\item \ctext{yellow!20}{\textbf{3. Resume the temporal co-denoising with DDIM using locally and globally refined output.}} 
\sname reverts the refined and fused $\mathbf{x}'_{0}$ to the previous timestep, producing $\mathbf{x}'_{t-1}$.
Here, the intermediate sample $\mathbf{x}'_{t}$ is re-derived by adding noise (from the second stage) to the $\mathbf{x}'_{0}$.
Finally, $\mathbf{x}'_{t-1}$ is computed using both $\mathbf{x}'_{0}$ and the re-derived $\mathbf{x}'_t$.

\end{itemize}

This one-step denoising process, comprising three stages, is iteratively repeated from $t=1000$ until $t=0$.
We present the complete algorithm in Algorithm~\ref{alg:main_alg}, with pseudo-code for each stage provided in Appendix~\ref{appen:pseudo}.

\input{resources/alg_sampling}

\subsection{\texorpdfstring{Align denoising paths across three stages\\for synchronized coupling}{Align denoising paths across three stages for synchronized coupling}}\label{method:component}

\sname introduces three stages to capture both local smoothness and global coherence.
However, synchronizing denoising trajectories across stages is crucial, as alternating between them can cause misalignment, leading to artifacts and reduced prompt fidelity.
In this section, we present the key components of each stage in \sname that ensure synchronized coupled sampling for seamless stage transitions.

\vspace{0.05in}
\noindent\textbf{Grounded timestep between the first and second stage.}
Aligning timesteps across stages is crucial in a three-stage process, as diffusion models establish the overall structure in earlier timesteps and refine finer details in later timesteps. If each stage operates on inconsistent temporal references, denoising trajectories become misaligned, introducing artifacts and inconsistencies in the final video (see Figure~\ref{fig:abl_component}).
To prevent this, \sname anchors the second-stage timestep to the first-stage sampling schedule, following the DDIM timestep progressing from $t=1000$ to $t=0$ based on designated sampling steps.
This ensures that both stages operate within a unified temporal reference, first establishing a coherent structure and then focusing on finer details throughout the denoising process.
This approach distinguishes \sname from standard CSD-based optimization (Equation~\ref{eqn:csd}), where timesteps are randomly selected at each iteration.

\vspace{0.05in}
\noindent\textbf{Fixed baseline noise for the second and third stage.}
In \sname, while ensuring global coherence, it is crucial in the second stage to preserve distinct prompt guidance for each chunk and prevent sample collapse, as discussed in Section~\ref{motivation:fusion_csd}.
To achieve this, we fix a single noise and use it consistently throughout one-step denoising, stabilizing second-stage optimization by providing a consistent baseline noise, unlike Equation~\ref{eqn:csd}, where random noise is introduced at each step.
Specifically, a fixed baseline noise $\bm{\epsilon}_{\text{base}}'$ is sampled from a Gaussian distribution at the start of the second stage, matching the shape of $\mathbf{x}'_0$. 
Each chunk $\mathbf{x}_{0}^{(i)}$ is then processed alongside its corresponding noise chunk $\bm{\epsilon}_{\text{base}}^{(i)}$. 
Additionally, \sname retains this fixed noise in the third stage, where it is reintroduced into the refined $\mathbf{x}_0$, further maintaining aligned update directions across stages.

\vspace{0.05in}
\noindent\textbf{Coupled sampling for the early timesteps.}
To balance local smoothness and global coherence, \sname regulates the second stage using $t_{\text{min}}$, applying it only until $t_{\text{min}} \in [800, 900]$, which we empirically found to be the optimal range. 
By this point, \sname establishes a coherent semantic layout across video chunks, preventing denoising trajectories from diverging as shown in Figure~\ref{fig:motiv_tsne}.
Thus, beyond $t_{\text{min}}$, \sname performs only the first and third stages, focusing on adding fine-grained details on an already structured layout.

\input{resources/tab_quan}

\vspace{0.05in}
\noindent\textbf{Structured prompt.} To further enhance coherent long scenarios with dynamics, \sname designs a structured prompt for multi-event scenarios, consisting of a shared global prompt for scenario-wide coherence and local prompts for event-specific variations (\eg, motions, camera movement, or attributes).
Na\"ively using a sequence of prompts (\eg, ``A teddy bear is standing'' $\rightarrow$ ``A teddy bear is running'') can lead to inconsistencies in the shared entity.
To reduce ambiguity and constrain the generation space, we introduce a detailed global prompt (\eg, ``A brown teddy bear with button eyes and a stitched smile''), ensuring uniformity across all video chunks.
Each chunk is then assigned a local prompt that builds on the global prompt, incorporating chunk-specific dynamics.
See Appendix~\ref{appen:impl_details} for details.

\vspace{0.05in}
\noindent\textbf{Flexibility of \sname to various T2V diffusion models.}
Notably, \sname is architecture-agnostic and compatible with any T2V diffusion model, supporting various diffusion objectives ($v$-prediction~\citep{salimans2022progressive}, $\epsilon$-prediction~\citep{guo2023animatediff, wang2023videofactory}), both diffusion-based and rectified flow-based models.
In addition, unlike prior works~\citep{tan2024video, cai2024ditctrl, qiu2023freenoise} restricted to U-Net~\citep{ronneberger2015u} or DiT~\citep{peebles2023scalable}, \sname remains flexible across architectures.
While we describe our method using $\epsilon$-prediction networks for clarity, \sname easily adapts to other diffusion settings by modifying $\mathcal{L}_{\tt{CSD}}$ in Algorithm~\ref{alg:main_alg} (see Appendix~\ref{appen:derivations} for a complete derivation on various diffusion settings).
We further validate this through comprehensive experiments on various denoising models in Section~\ref{sec:exp} and Appendix~\ref{appen:additional_abl}.

%% file: resources/alg_sampling.tex
\begin{algorithm}[t]
\begin{spacing}{1}
\SetCommentSty{small}

\caption{\\ \lname (\sname)}\label{alg:main_alg}

\small \SetKwFunction{DDIM}{DDIM}
\SetKwFunction{DiffusionForward}{DiffForward}
\SetKwFunction{TakeChunk}{TakeChunk}
\SetKwFunction{Sync}{Fusion}
\SetKwFunction{AddNoise}{AddNoise}
\SetKwProg{Fn}{Procedure}{:}{}
\SetKwInOut{Require}{Require}
\SetKwInOut{Ensure}{Ensure}

\Require{
$\Phi$\tcp*{a pre-trained T2V model} 
$\{\mathbf{c}^{(i)}\}_{i=1}^N$\tcp*{conditioning text-prompt embeddings}
}
\Ensure{$\mathbf{x}'_{0}$}

\Fn{Ours}{
    $\mathbf{x}'_{T} \sim \mathcal{N}(\mathbf{0},\mathbf{I})$\;
    
    \For{$t = T, ..., 1$}{

    \boxit{green!20}{5.3}
        \For{$i = 0, ..., N-1$}{
            $\mathbf{x}_t^{(i)} \gets \TakeChunk(\mathbf{x}'_t)$\;
            $\bm{\epsilon}_{\text{pred}}^{(i)} \gets \DiffusionForward_{\Phi}(\mathbf{x}_t^{(i)}, \mathbf{c}^{(i)}, t)$\;
            ${\hat{\mathbf{x}}_{0|t}}^{(i)} \gets \DDIM(\bm{\epsilon}_{\text{pred}}^{(i)}, t)$\;
        }
        
        $\mathbf{x}'_0 \gets \Sync(\{ \hat{\mathbf{x}}_{0|t}^{(i)}\}_{i=1}^N)$\;

        \boxit{pink}{17.4}
        \If{$t > t_{\text{min}}$}{

            $\bm{\epsilon}'_{\text{base}} \sim \mathcal{N}(\mathbf{0},\mathbf{I})$\tcp*{fixed baseline noise}

            \For{$\text{iter} = 1$ \KwTo $\text{iters}$}{

            \For{$i = 0, ..., N-1$}{
                $\mathbf{x}_0^{(i)} \gets \TakeChunk(\mathbf{x}'_0)$\;

                $\bm{\epsilon}_{\text{base}}^{(i)} \gets \TakeChunk(\bm{\epsilon}'_{\text{base}})$\;

                $\mathbf{x}_t^{(i)} \gets \AddNoise(\mathbf{x}_0^{(i)}, \bm{\epsilon}_{\text{base}}^{(i)})$\;

                $\bm{\epsilon}_{\text{pred}}^{(i)} \gets \DiffusionForward_{\Phi}(\mathbf{x}_t^{(i)}, \mathbf{c}^{(i)}, t)$\tcp*{grounded timestep}
            }
            
            \For{$i = 0, ..., N-1$}{
                $\nabla_{\mathbf{x}_0^{(i)}} \mathcal{L}_{\tt{CSD}}(\mathbf{x}_t^{(i)}, \bm{\epsilon}_{\text{base}}^{(i)}, \bm{\epsilon}_{\text{pred}}^{(i)})$\;
            }
            $\nabla_{\mathbf{x}'_0} \mathcal{L}_{\tt{CSD}} \gets \Sync(\{\nabla_{\mathbf{x}_0^{(i)}} \mathcal{L}_{\tt{CSD}}\}_{i=1}^N)$\;
            Backpropagate $\nabla_{\mathbf{x}'_0} \mathcal{L}_{\tt{CSD}}$\;
            SGD update on $\mathbf{x}'_0$\;
        }
        }
        
        \boxit{yellow}{0.4}
        $\mathbf{x}'_{t-1} \gets \DDIM(\mathbf{x}'_0, \bm{\epsilon}_{\text{base}}, t)$\;
    }
}
\end{spacing}
\end{algorithm}

%% file: resources/tab_quan.tex
\begin{figure*}
\vspace{-0.1in}  
\captionof{table}{
\textbf{Quantitative comparison.} 
Experimental results of \sname with baselines on multi-event long video generations.
\textbf{Bold} indicates the best results.
\sname consistently outperforms baselines across temporal consistency, per-frame quality, and prompt fidelity.
}
\vspace{-0.13in}
\label{tab:main}
\centering
\resizebox{0.95\textwidth}{!}{
\begin{tabular}{c l ccccc cc cc}
\toprule
    & & \multicolumn{5}{c}{Temporal Quality}  & \multicolumn{2}{c}{Frame-wise Quality} & \multicolumn{2}{c}{Semantics} \\
    \cmidrule(lr){3-7} \cmidrule(lr){8-9}  \cmidrule(lr){10-11} 
    & & \multicolumn{1}{c}{Subject} & \multicolumn{1}{c}{Background} & \multicolumn{1}{c}{Motion} & \multicolumn{1}{c}{Dynamic} & \multicolumn{1}{c}{Num} & \multicolumn{1}{c}{Aesthetic} & \multicolumn{1}{c}{Imaging}  & \multicolumn{1}{c}{Glb Prompt} & \multicolumn{1}{c}{Loc Prompt}  \\
    Backbone & Method & \multicolumn{1}{c}{Consistency $\uparrow$} & \multicolumn{1}{c}{Consistency $\uparrow$} & \multicolumn{1}{c}{Smoothness $\uparrow$} & \multicolumn{1}{c}{Degree $\uparrow$} & \multicolumn{1}{c}{Scenes $\downarrow$} & \multicolumn{1}{c}{Quality $\uparrow$} & \multicolumn{1}{c}{Quality $\uparrow$} & \multicolumn{1}{c}{Fidelity $\uparrow$} & \multicolumn{1}{c}{Fidelity $\uparrow$} \\
    \midrule
      \multirow{3.5}{*}{CogVideoX~\citep{yang2024cogvideox}} 
            & Gen-L-Video~\citep{wang2023gen}
            & 81.34\% & 89.46\% & \textbf{98.32}\% & 50.00\% & 2.292 & 60.09\% & 58.52\% & 0.324 & 0.351 \\
            & FIFO-Diffusion~\citep{kim2024fifo}
            & 83.54\% & 90.72\% & 98.04\% & 70.83\% & 1.938 & 59.59\% & 63.18\% & 0.323 & 0.327\\
            \cmidrule{2-11}
            & \textbf{\sname~(Ours)}
            & \textbf{88.92}\% & \textbf{94.57}\% & 98.21\% & \textbf{85.42}\% & \textbf{1.208} & \textbf{63.37}\% & \textbf{67.43}\%  & \textbf{0.341} & \textbf{0.354} \\

    \midrule

    \multirow{3.5}{*}{Open-Sora Plan~\citep{lin2024open}} 
            & Gen-L-Video~\citep{wang2023gen}
            & 85.15\% & 92.63\% & 98.93\% & 43.75\% & 2.458 & 61.44\%  & 57.19\% & 0.319 & 0.337\\
            & FIFO-Diffusion~\citep{kim2024fifo}
            & 89.14\% & 94.34\% & 98.19\% & 27.08\% & \textbf{1.125} & 60.26\% & \textbf{61.23}\% & 0.327 & 0.331\\
            \cmidrule{2-11}
            & \textbf{\sname~(Ours)}
            & \textbf{90.19}\% & \textbf{94.78}\% & \textbf{99.06}\% & \textbf{58.33}\% & 1.646 & \textbf{63.79}\% & 60.19\% & \textbf{0.328} & \textbf{0.345}\\
    \bottomrule
\end{tabular}
}

\vspace{0.02in}

\includegraphics[width=0.98\textwidth]{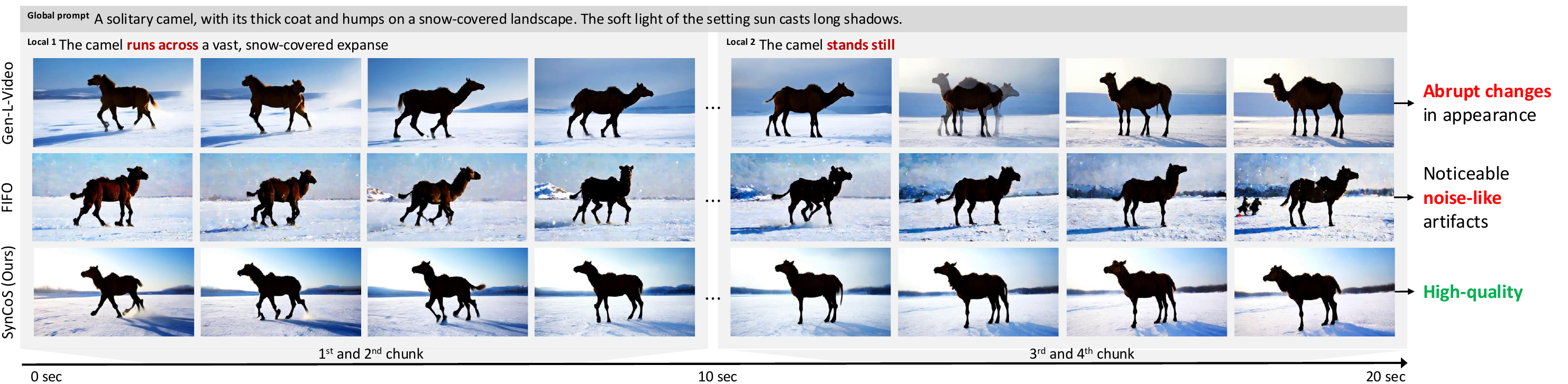}
\vspace{-0.15in}
\captionof{figure}{\textbf{Qualitative comparisons} on Open-Sora Plan~\citep{lin2024open}. 
All examples are 4 times longer in duration compared to the underlying base model, generating a 20-second video. Gen-L-Video~\citep{wang2023gen} suffers from abrupt appearance changes, while FIFO-Diffusion~\citep{kim2024fifo} introduces noticeable noise-like artifacts.
In contrast, our proposed method, \sname, generates high-quality, temporally coherent videos that faithfully follow the prompt throughout the sequence.
}\label{fig:main_qual}
\vspace{-0.21in}

\end{figure*}

%% file: sec/6_exp.tex
\section{Experiments}\label{sec:exp}
We first provide a brief overview of the experimental setup in Section~\ref{sec:exp_setup}, followed by a thorough validation of our inference framework across diverse scenarios and diffusion settings in Section~\ref{sec:exp_main}, Appendix~\ref{appen:disc_baselines}, comparing it with previous works.
Additionally, we conduct comprehensive ablation studies in Section~\ref{sec:exp_abl} and Appendix~\ref{appen:additional_abl}.

\subsection{Experimental setup}\label{sec:exp_setup}

\noindent\textbf{Implementation details.}
We evaluate on 48 long-video scenarios, forming a more extensive benchmark than previous works~\citep{kim2024fifo, oh2023mtvg}, with video lengths extended by 4–5 times.
These multi-event scenarios encompass challenges such as object motion control, camera control, background changes, compositional generation, and physical transformations (full prompt details in Appendix~\ref{appen:impl_details}).

To demonstrate broad applicability, we implement \sname across four different T2V diffusion models: CogVideoX-2B~\citep{yang2024cogvideox} and Open-Sora Plan (v1.3)~\citep{lin2024open} (in Section~\ref{sec:exp_main}) as well as a modified Open-Sora Plan (v1.2) variant (denoted as M) and VideoCrafter2~\citep{wang2023videofactory} (in Appendix~\ref{appen:additional_abl} and~\ref{appen:disc_baselines}).
All experiments are conducted on a single NVIDIA H100 80GB GPU, with inference time measurements detailed in Appendix~\ref{appen:impl_details}.

\vspace{0.05in}
\noindent\textbf{Baselines.}
We primarily evaluate \sname against tuning-free, architecture-agnostic baselines (Gen-L-Video~\citep{wang2023gen}, FIFO-Diffusion~\citep{kim2024fifo}) and provide additional comparisons with architecture-specific methods (FreeNoise~\citep{qiu2023freenoise}, Video-Infinity~\citep{tan2024video}, DiTCtrl~\citep{cai2024ditctrl}) in Appendix~\ref{appen:disc_baselines} for a fair evaluation within their respective architectures.
For all baselines, we follow their reported experimental setups and carefully tune hyperparameters for optimal performance.

\vspace{0.05in}
\noindent\textbf{Evaluation metrics.}
We evaluate multi-event long video generation results using VBench~\citep{huang2024vbench} to assess both temporal consistency and per-frame quality.
Temporal consistency is measured by subject and background consistency, motion smoothness, and dynamic degree, while aesthetic and imaging quality evaluate per-frame fidelity.
Following prior works~\citep{henschel2024streamingt2v, xie2024progressive}, we use Adaptive Detector~\citep{Castellano_PySceneDetect} to count scene changes, where a `Num Scenes' value of 1 indicates no transitions.
To assess prompt-driven controllability, we measure prompt fidelity using CLIP~\citep{radford2021learning}, computing image-text similarity for both local and global prompts.

\input{resources/abl_components}
\subsection{Main experiments}\label{sec:exp_main}
While \sname incorporates a structured prompt, we apply the structured prompt to all baselines for a fair comparison.

\vspace{0.05in}
\noindent\textbf{Quantitative comparisons.}
In Table~\ref{tab:main}, \sname significantly outperforms existing tuning-free baselines for long video generation, achieving higher scores in both temporal consistency and per-frame image quality.
In particular, \sname surpasses baselines in subject and background consistency while achieving the highest dynamic degree, demonstrating its ability to generate visually dynamic but temporally consistent videos.
Since a lower dynamic degree naturally improves consistency, high-scores of \sname in both metrics are particularly significant.
Additionally, \sname demonstrates superior prompt fidelity, generating distinct prompt-controlled dynamics while maintaining semantic and content consistency across the entire video.

\vspace{0.05in}
\noindent\textbf{Qualitative comparisons.}
As shown in Figure~\ref{fig:emph} and Figure~\ref{fig:main_qual}, \sname excels in high-quality multi-event long video generation, ensuring temporal consistency throughout the video.
Gen-L-Video suffers from overlapping artifacts due to diverging denoising paths across video chunks when guided by different prompts.
\sname effectively mitigates these issues through global synchronization.
FIFO-Diffusion exhibits style drift on CogVideoX-2B and suffers from degraded image quality with noise artifacts on Open-Sora Plan.
This issue arises from its sequential timestep processing of adjacent frames, leading to a training-inference discrepancy.
As Open-Sora Plan encodes more frames per video chunk, FIFO-Diffusion struggles with severe timestep variations between frames, resulting in pronounced artifacts.
In contrast, \sname does not introduce any training-inference discrepancies.
See Appendix~\ref{appen:disc_baselines} for further comparisons on baselines and Appendix~\ref{appen:qual} for additional qualitative results.

\subsection{Ablation study}\label{sec:exp_abl}
Table~\ref{tab:comp_abl} and Figure~\ref{fig:abl_component} ablate the key components essential for coupling the three stages into a unified sampling, providing both quantitative and qualitative evaluations.

First, grounded timesteps anchor the generation process, ensuring that each stage focuses on the same aspects of generation when alternating between stages.
Without grounding, random timesteps disrupt the progressive generation from structure formation (early timesteps) to detail addition (later timesteps), causing content drift (see w/o grounded timestep in Table~\ref{tab:comp_abl}).
For example, in Figure~\ref{fig:abl_component}, the absence of grounded timesteps causes a volcano to randomly change its appearance, generating one or two peaks across frames due to misalignment between structure and detailing.

Additionally, fixed baseline noise is crucial for preserving distinct prompt guidance while ensuring cohesion between them (see w/o fixed baseline noise in Table~\ref{tab:comp_abl}).
Without it, distinct prompt guidance weakens as cohesion dominates, reducing prompt fidelity (\eg, a volcano without an eruption or missing rising smoke, as shown in Figure~\ref{fig:abl_component}).
Moreover, $t_{\text{min}}$ regulates the synchronization strength (see w/o $t_{\text{min}}$ in Table~\ref{tab:comp_abl}).
Lastly, structured prompts improve coherence by blending global prompts for consistency with local prompts for fine-grained control (see w/o structured prompt in Table~\ref{tab:comp_abl}).
In addition to qualitative examples in Section~\ref{sec:motivation}, Appendix~\ref{appen:additional_abl} presents quantitative ablations by skipping each stage of \sname to assess the efficacy of coupled stages.

%% file: resources/abl_components.tex
\begin{figure*}[ht]
\centering\small

    \vspace{-0.1in}
    
    \begin{minipage}{0.33\textwidth}
        \centering
        \captionof{table}{\textbf{Quantitative ablation study}. *Abbreviations: subject consistency (SC), background consistency (BC), aesthetic quality (AQ), and prompt fidelity (PF).}
        \label{tab:comp_abl}
        \vspace{-0.1in}
        \resizebox{\linewidth}{!}{
        \begin{tabular}{l c c c c}
        \toprule
         & \multicolumn{2}{c}{Temporal}  & \multicolumn{1}{c}{Frame} & \multicolumn{1}{c}{Semantics} \\
        \cmidrule(lr){2-3} \cmidrule(lr){4-4}  \cmidrule(lr){5-5} 
        
         & SC $\uparrow$  & BC $\uparrow$  & AQ $\uparrow$  & PF $\uparrow$ \\
        \midrule
        SynCoS & 0.864 & 0.927 & 0.643 & 0.381 \\
        \midrule
        SynCoS w/o $t_{\text{min}}$ & 0.724 & 0.854 & 0.632 & 0.373 \\
        SynCoS w/o grounded timestep & 0.803 & 0.899 & 0.638 & 0.364\\
        SynCoS w/o fixed baseline noise & 0.817 & 0.904 & 0.643 & 0.382 \\
        SynCoS w/o structured prompt & 0.837 & 0.903 & 0.663 & 0.362 \\
        \bottomrule
        \end{tabular}
        }
    \end{minipage}
    \hfill
    \begin{minipage}{0.66\textwidth}
        \centering
        \includegraphics[width=0.92\textwidth]{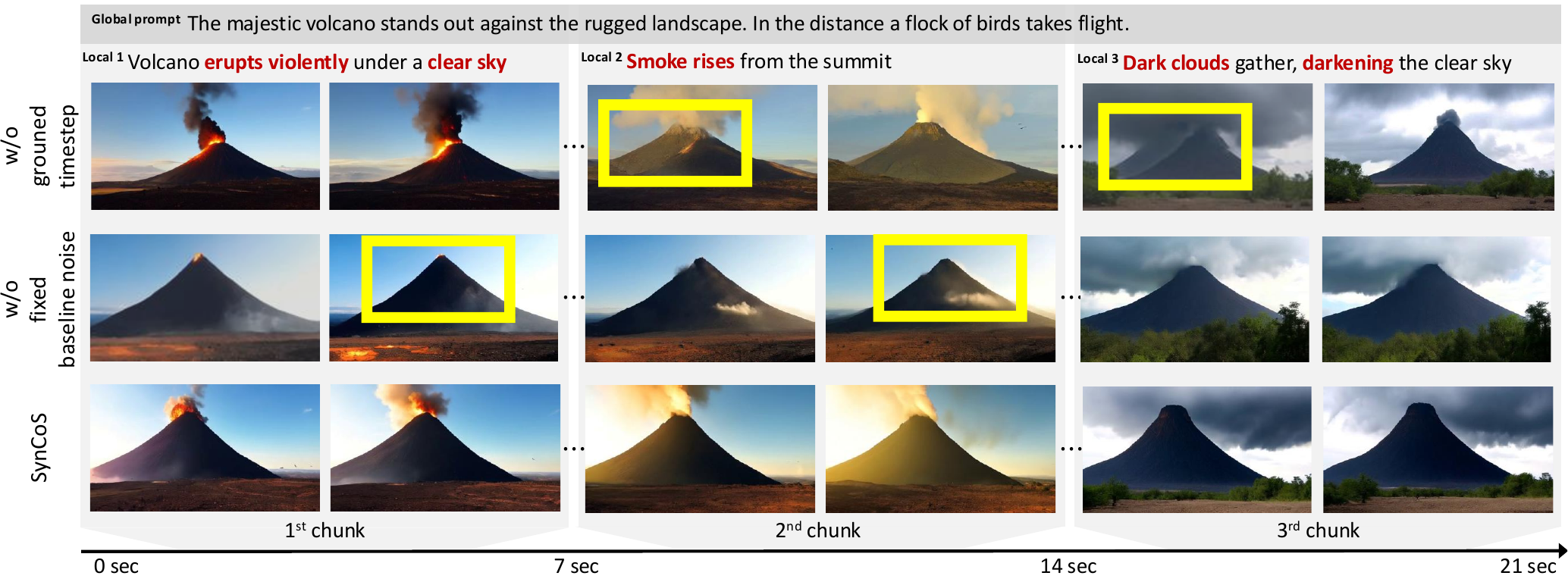}
        \vspace{-0.15in}
        \caption{\textbf{Qualitative ablation study}. Without a grounded timestep, structural inconsistencies arise (\eg, the volcano alternates between one and two peaks).
        Without a fixed baseline noise, it fails to follow local prompts faithfully (\eg, missing eruptions or absent smoke).
        }
        \label{fig:abl_component}
    \end{minipage}

    \vspace{-0.2in}
    
\end{figure*}

%% file: sec/2_related.tex
\section{Related work}\label{sec:related}
\noindent\textbf{Text-guided video diffusion models.}
The success of text-to-image (T2I) generation with diffusion models~\citep{rombach2021highresolution,saharia2022photorealistic,balaji2022ediffi} has inspired extensions to the more complex task of text-to-video generation.
Several works extended T2I models by fine-tuning them with temporal layers, adapting spatial structures to temporal dynamics for more efficient training~\citep{ho2022imagen,wang2023videofactory,an2023latent,blattmann2023align,ge2023preserve,he2022lvdm,singer2022make}.
Recent advancements in Diffusion Transformers (DiT)~\citep{peebles2023scalable} and flow-based generative models~\citep{liu2022flow} have further improved quality, enabling state-of-the-art T2V diffusion models~\citep{jin2024pyramidal}. 
However, high computational and memory costs restrict these models to short video clips, limiting their applicability to long video generation.

\vspace{0.05in}
\noindent\textbf{Training-based long video generation.}
To extend video length, alternative strategies include different architecture choices: transformer-based methods, autoregressive diffusion models, and hierarchical models~\citep{villegas2022phenaki, ge2022long, xie2024progressive, henschel2024streamingt2v, tian2024videotetris, yin2023nuwa, chen2023seine}.
Transformer-based methods~\citep{villegas2022phenaki, ge2022long} generate long videos in a one-shot manner from multiple prompts. However, they require extensive training from scratch and often suffer from content drift, degrading quality over time.
Autoregressive diffusion models~\citep{henschel2024streamingt2v, xie2024progressive} generate frames sequentially, conditioning each chunk on the previous one. 
While this improves short-term consistency, autoregressive dependencies cause errors to accumulate, resulting in visual artifacts and inconsistencies in long-form synthesis. Additionally, these methods are limited to single-prompt generation, making them unsuitable for evolving content.
Alternatively, hierarchical diffusion models~\citep{chen2023seine, yin2023nuwa} generate keyframes first and then interpolate intermediate frames. 
While this approach maintains structural consistency, interpolation alone does not introduce new content, limiting dynamic scene evolution.

\vspace{0.05in}
\noindent\textbf{Tuning-free long video generation.}
Recent approaches extend existing T2V diffusion models for long video generation in a tuning-free manner using multiple prompts~\citep{wang2023gen, kim2024fifo, oh2023mtvg, qiu2023freenoise}.
While scalable without additional training, these approaches focus on local frame transitions, often leading to content drift and semantic inconsistencies over longer sequences.
Gen-L-Video \citep{wang2023gen} fuses denoising paths in overlapping frames, but local fusion dilutes prompt guidance from different prompts between frames, degrading quality and causing divergent denoising paths.
FIFO-Diffusion~\citep{kim2024fifo} enables infinite sampling by sequentially assigning timesteps, but discrepancies between training and inference lead to undesirable artifacts.
FreeNoise~\citep{qiu2023freenoise} uses window-based attention fusion to attend to longer frame dependencies, but it is incompatible with newer diffusion models that lack separate spatial and temporal attention layers.
In contrast, our method synchronizes local and global denoising paths across adjacent and distant frames, preventing content drift and preserving semantic consistency throughout long videos.
Additionally, Video-Infinity~\citep{tan2024video} and DiTCtrl~\citep{cai2024ditctrl} propose architecture-specific extensions, whereas our approach is compatible with any T2V diffusion model.
Please refer to Appendix~\ref{appen:disc_baselines} for further discussion and comparisons.

%% file: sec/7_conclusion.tex
\section{Conclusion}\label{sec:conclusion}
This work presents a tuning-free inference framework that extends any existing T2V diffusion model for multi-event long video generation.
Unlike previous works focusing solely on local smoothing between adjacent frames, our approach simultaneously ensures smooth local transitions and global coherence by introducing three synchronously coupled stages with a structured prompt.
Extensive evaluations across diverse, challenging, long video scenarios with multiple events demonstrate that our method generates high-quality, temporally coherent long videos, significantly outperforming prior works both quantitatively and qualitatively.

%% file: sec/X_suppl.tex
\clearpage
\setcounter{page}{1}

\maketitlesupplementary

{\small{
\textbf{Project page:} \url{https://syncos2025.github.io/}
}
}

\input{appen/derivations}

\input{appen/exp_details}

\input{appen/additional_abl}

\input{appen/disc_baselines}

\input{appen/pseudo_code}

\input{appen/qual_comp_additional}

%% file: appen/derivations.tex
\section{Foundational concepts and derivations}\label{appen:derivations}

\subsection{Further details on preliminaries}
This section provides additional details on the preliminaries introduced in Section~\ref{sec:pre} to further guide the readers.

\vspace{0.05in}
\noindent\textbf{Diffusion models.}
Diffusion models are generative models that learn to gradually transform a noise sample from a tractable noise distribution towards a target data distribution. This transformation is consisting of two processes: a forward process and a reverse process.
In the forward process, noise is incrementally added to the data sample over a sequence of timestep $t$, leading to a gradual loss of structure in the data. This forward process is defined as following:
\begin{equation}
    \mathbf{x}_t = \sqrt{\bar\alpha_t} \mathbf{x}_0 + \sqrt{1-\bar\alpha_t} \bm{\epsilon}
\end{equation}\label{eqn:diff_forward}
where $\bar\alpha_t$ is a pre-defined coefficient that regulates noise schedule over time, and $\bm{\epsilon} \sim \mathcal{N}(\mathbf{0},\mathbf{I})$ denotes a noise sampled from a standard normal distribution.
The reverse diffusion process aims to invert the forward process, gradually removing the noise and recovering the original target data distribution.
This reverse transformation is modeled using a neural network (referred to as the diffusion model, $\bm{\epsilon}_\Phi$), which is trained using a loss function based on denoising score matching~\cite{ho2020denoising, song2020score}. The loss function is defined as:

\begin{equation}
\begin{aligned}
    \mathcal{L}_{\tt{diff}}(\Phi; \mathbf{x}) = \mathbb{E}_{t, \bm{\epsilon} \sim \mathcal{N}(\mathbf{0},\mathbf{I})} 
    \left[ w(t) \|\bm{\epsilon}_\Phi(\mathbf{x}_t; t) - \bm{\epsilon}\|^2_2 \right],
\end{aligned}\label{eqn:diff_obj}
\end{equation}

where $w(t)$ is a weighting function applied to each timestep $t$, and $t \sim \mathcal{U}(0, 1)$ is drawn from a uniform distribution.


\vspace{0.05in}
\noindent\textbf{Score distillation sampling.} Score distillation sampling (SDS)~\citep{poole2022dreamfusion} introduces a new way to optimize any arbitrary differentiable parametric function $g$ using diffusion models as a critic by posing generative sampling as an optimization problem.
The flexibility of SDS in optimizing diverse differentiable operators has made it a versatile tool for visual tasks.
Specifically, given $\mathbf{x} = g(\theta)$, the gradient of the diffusion loss function of Equation~\ref{eqn:diff_obj} with respect to the parameter $\theta$ is expressed as below:

{\small
\begin{equation}
\begin{aligned} 
\nabla_{\theta} & \mathcal{L}_{\tt{SDS}} 
 (\Phi; \mathbf{x} = g(\theta) ) 
\\
& \stackrel{\Delta}{=} \mathbb{E}_{t, \bm{\epsilon}} 
\left[ w(t)\left(\bm{\epsilon}_\Phi^\gamma(\mathbf{x}_t; \mathbf{c}, t) - \bm{\epsilon}
\right) 
\frac{\partial \bm{\epsilon}_\Phi^\gamma(\mathbf{x}_t; \mathbf{c}, t)}{\partial \mathbf{x} }
\frac{\partial \mathbf{x}}{\partial \theta } \right]
\end{aligned}\label{eqn:full_sds}
\end{equation}
}

In DreamFusion~\citep{poole2022dreamfusion}, it has been shown that by omitting the Jacobian term of the U-Net (the middle term) in Equation~\ref{eqn:full_sds} yields an effective gradient for estimating an update direction that follows the score function of diffusion models, thereby moving towards a higher-density region. 
The simplified expression is given as Equation~\ref{eqn:sds} in Section~\ref{sec:pre}.


\vspace{0.05in}
\noindent\textbf{Collaborative score distillation sampling.} The original SDS method considers only a single sample, $\mathbf{x}$, during the optimization process.
Collaborative Score Distillation (CSD)~\citep{kim2023collaborative} 
extends SDS to enable inter-sample consistency by synchronizing gradient updates across multiple samples, $\{\mathbf{x}^{(i)}\}_{i=1}^N$.
Specifically, CSD generalizes SDS by using Stein Variational Gradient Descent (SVGD)~\citep{liu2016stein} to align gradient directions across multiple samples as Equation~\ref{eqn:csd} in Section~\ref{sec:pre}.
For a positive definite kernel, $k$, in Equation~\ref{eqn:csd}, CSD employs the standard Radial Basis Function (RBF) kernel.
This approach ensures that each parameter update is influenced by other samples with scores adjusted via importance weights based on sample affinity, thereby effectively promoting inter-sample consistency during optimization.

\input{appen/resources/prompt}

\vspace{0.05in}
\subsection{\sname with different diffusion models}
In this paper, we present our method using diffusion models with $\epsilon$-prediction networks for clarity.
However, our framework is compatible with any T2V denoising model.
To illustrate this, we describe how applying our framework to different diffusion settings.

\noindent\textbf{Flow-based models.}
Recently, rectified flow~\citep{lipman2022flow}, a unified ODE-based framework for generative modeling, introduces a simplified denoising process that optimizes the trajectories in diffusion space to be as straight as possible. 
Given a data sample $\mathbf{x}$ from the data distribution and a noise sample $\bm{\epsilon}$ from a Gaussian distribution, rectified flow defines the forward process as:
\begin{equation}
\mathbf{x}_t = t\bm{\epsilon} + (1-t)\mathbf{x}_0, \quad t \in [0,1]
\end{equation}

Accordingly, the reverse process is governed by an ODE that maps $\bm{\epsilon}$ to $\mathbf{x}_0$:
\begin{equation}
d\mathbf{x}_t = \mathbf{v}(\mathbf{x}_t; t) dt, \quad t \in [0,1]
\end{equation}
where $\mathbf{v}$ is a velocity field estimated by a learnable neural network $\Phi$, where the model is trained using the following objective:
{\small
\begin{equation}
\mathcal{L}_{\tt{flow}}(\Phi; \mathbf{x}) = \mathbb{E}_{t, \bm{\epsilon}} \left[ w(t) \| (\bm{\epsilon} - \mathbf{x}_{0}) - \mathbf{v}(\mathbf{x}_t; t) \|_2^2 \right].
\label{eq:training_loss}
\end{equation}
}

\vspace{0.05in}
\noindent\textbf{Score distillation with flow-based models.}
While SDS and CSD have been implemented on denoising diffusion models with $\epsilon$-prediction networks, the core concept of score distillation—using diffusion models as generative priors for optimization—can also be extended to flow-based models.
Below, we derive the equation for score distillation adapted to flow-based models.
By computing the gradient of the training objective $\mathcal{L}_{\tt{flow}}$ for flow-based models, $\Phi$, with respect to $\theta$, the score distillation sampling adapted to flow-based models can be expressed as:
{\small
\begin{equation}
\begin{aligned} 
\nabla_{\theta_{i}} & \mathcal{L}_{\tt{Flow-SDS}} 
 (\Phi, \mathbf{x} = g(\theta) ) 
\\
& \stackrel{\Delta}{=} \mathbb{E}_{\bm{\epsilon}, t} 
\left[ w(t)\left(\bm{v}_\Phi(\mathbf{x}_{t}, t) - \left(\bm{\epsilon} - \mathbf{x} \right) \right) \frac{\partial 
\mathbf{x}}{\partial \theta } \right]. 
\end{aligned}
\end{equation}
}

Here, $w(t)$ is a time-dependent weighting function, and $\bm{v}_{\Phi}$ is estimated by the pre-trained flow-based denoising network $\Phi$. 
In accordance with SDS conventions, the (transformer) Jacobian term is omitted to enhance computational efficiency, enabling the optimization of $\mathbf{x}$ using the rectified flow model. 
This loss is then could be reinterpreted within the Collaborative Score Distillation (CSD) framework~\citep{kim2023collaborative}, allowing for synchronous updates across multiple samples instead of treating each gradient independently.

\definecolor{background}{rgb}{0.95, 0.95, 0.95}
\definecolor{keyword}{rgb}{0.0, 0.0, 0.6}
\definecolor{comment}{rgb}{0.0, 0.5, 0.0}
\definecolor{string}{rgb}{0.58, 0.0, 0.82}





%% file: appen/resources/prompt.tex
\begin{figure*}[ht]
\centering\small

\includegraphics[width=0.9\textwidth]{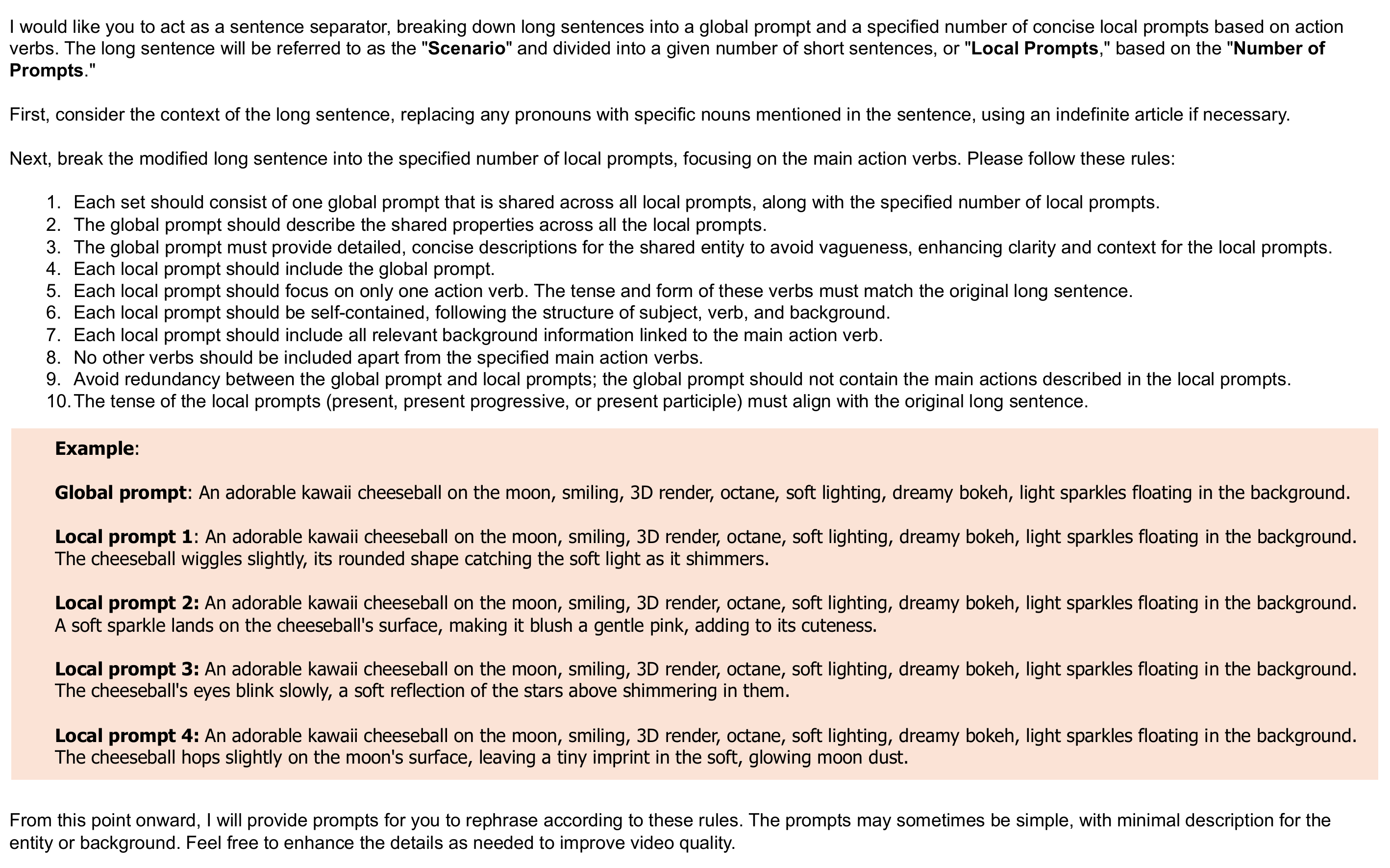}

\vspace{-0.02in}
\caption{\textbf{Instruction for generating structured prompt}. This instruction follows the guidelines to create individual local prompts and a shared global prompt based on a scenario and the number of prompts the user gives.}\label{fig:appen_prompt}
\end{figure*}

%% file: appen/exp_details.tex
\input{resources/main_table2}
\section{Experimental details}\label{appen:impl_details}

\noindent\textbf{Details of structured prompt.}\label{appen:prompt}
The structured prompt is designed to divide a cohesive story (\ie, scenario) into multiple prompts, enabling the generation of long videos controlled by distinct prompts. It consists of a single global prompt that describes shared properties across all local prompts (e.g., object appearances or styles) and distinct local prompts that specify changes in objects, motions, or styles.
To create a structured prompt, we leverage a Large Language Model (LLM), as in previous works~\citep{oh2023mtvg}. The process involves instructing the LLM to segment a given scenario into multiple local prompts and a detailed global prompt, as illustrated in Figure~\ref{fig:appen_prompt}. 
We use GPT-4o~\citep{gpt-4o} to generate these structured prompts, ensuring consistency across all experiments.
These structured prompts are applied in all experiments, including the main evaluations of our method and the baselines, as well as in ablation studies, ensuring a fair comparison.

To extensively verify our method under various challenging scenarios, we consider both previously established scenarios~\citep{qiu2023freenoise, oh2023mtvg, tian2024videotetris, lu2025freelong, wang2023gen, kim2024fifo} and our new, more challenging scenarios, including: background changes, object motion changes, camera movements, compositional generation, complex scene transitions, cinematic effects, physical transformation, and storytelling.
We experiment with 48 structured prompt scenarios, including 11 with two local prompts, 12 with three, and 25 with four, extending video length by a factor of four or five.
Full lists are provided on our project page: \url{https://syncos2025.github.io/}

\vspace{0.05in}
\noindent\textbf{Implementation details of the main experiments.} DDIM sampling was performed with 50 steps, setting DDIM $\eta$ to 0.
The stride $s$ is set to 4 for CogVideoX-2B and 6 for Open-Sora Plan (v1.3).
In second stage, we set $t_{\text{min}} \in [800, 900]$, learning rate, $lr \in [0.5, 1]$ using AdamW Optimizer~\citep{loshchilov2017decoupled}, and $iters \in [20, 50]$.
The scale of the classifier-free guidance is set to 6 for CogVideoX-2B and 7.5 for Open-Sora Plan (v1.3).
For Gen-L-Video, we use the same strides and guidance scale as in our experiments. 
For FIFO-Diffusion, we set $n = 4$ for the number of partitions in latent partitioning and lookahead denoising, following their best parameter configurations.

\vspace{0.05in}
\noindent\textbf{Measurements of computation time}.
We measure the computation time required to generate a $4\times$ longer video compared to the underlying base model of CogVideoX-2B, using a single H100 80GB GPU (Table~\ref{tab:comp_time}), including both main baselines and our approach.
Although our method takes $2.6\times$ longer than the baselines, it achieves notable gains in quality and consistency, demonstrating superior qualitative and quantitative performance.

While computation time is not our primary focus, it can be adjusted based on video scenarios and the required level of synchronization. Specifically, reducing $iters$ in the second-stage optimization can lower computation time.
Nonetheless, we think reducing computation time while maintaining quality is a promising direction in the future.

\begin{table}[ht]
\centering
\caption{Measurements and comparisons of computation time on CogVideoX-2B.}
\label{tab:comp_time}
\vspace{-0.1in}
\resizebox{0.8\linewidth}{!}{
\begin{tabular}{c c c}
    \toprule
    Gen-L-Video~\citep{wang2023gen} & FIFO~\citep{kim2024fifo} & SynCoS (Ours) \\
    \midrule
    21 min. & 21 min. & 55 min. \\
    \bottomrule
\end{tabular}
}
\end{table}

%% file: resources/main_table2.tex
\begin{figure*}[ht]
\captionof{table}{
\textbf{Quantitative ablations study of the three coupled stages in \sname}, omitting each stage during one-timestep denoising, demonstrates the importance of all three stages for coherent long video generation with multiple events.
}
\vspace{-0.05in}
\label{tab:abl_stages}
\centering
\resizebox{\textwidth}{!}{

\begin{tabular}{c ccc ccccc cc cc}
\toprule
    & & & & \multicolumn{5}{c}{Temporal Quality}  & \multicolumn{2}{c}{Frame-wise Quality} & \multicolumn{2}{c}{Semantics} \\
    \cmidrule(lr){5-9} \cmidrule(lr){10-11}  \cmidrule(lr){12-13} 
    
    &  \multicolumn{3}{c}{Stage} & \multicolumn{1}{c}{Subject} & \multicolumn{1}{c}{Background} & \multicolumn{1}{c}{Motion} & \multicolumn{1}{c}{Dynamic} & \multicolumn{1}{c}{Num} & \multicolumn{1}{c}{Aesthetic} & \multicolumn{1}{c}{Imaging}  & \multicolumn{1}{c}{Glb Prompt} & \multicolumn{1}{c}{Loc Prompt}  \\
    
    Backbone & 1 & 2 & 3 & \multicolumn{1}{c}{Consistency $\uparrow$} & \multicolumn{1}{c}{Consistency $\uparrow$} & \multicolumn{1}{c}{Smoothness $\uparrow$} & \multicolumn{1}{c}{Degree $\uparrow$} & \multicolumn{1}{c}{Scenes $\downarrow$} & \multicolumn{1}{c}{Quality $\uparrow$} & \multicolumn{1}{c}{Quality $\uparrow$} & \multicolumn{1}{c}{Fidelity $\uparrow$} & \multicolumn{1}{c}{Fidelity $\uparrow$} \\
    \midrule

      \multirow{3.5}{*}{M} 
            &\ck & \xk & \ck
            & 80.46\% & 91.14\% & \textbf{98.55}\% & 97.92\% & 1.229 & 53.36\% & 58.42\% & 0.318 & \textbf{0.348} \\
            &\xk & \ck & \xk
            & 78.88\% & 91.63\% & 97.70\% & 14.58\% & 21.33 & 45.56\% & 42.27\% & 0.305 & 0.300 \\
            \cmidrule{2-13}
            &\ck & \ck & \ck
            & \textbf{82.70}\% & \textbf{91.85}\% & 98.24\% & \textbf{100.00}\% & \textbf{1.042} & \textbf{54.56}\% & \textbf{65.53}\% & \textbf{0.325} & \textbf{0.348} \\

    \bottomrule
\end{tabular}
}

\vspace{0.1in}

\includegraphics[width=1\textwidth]{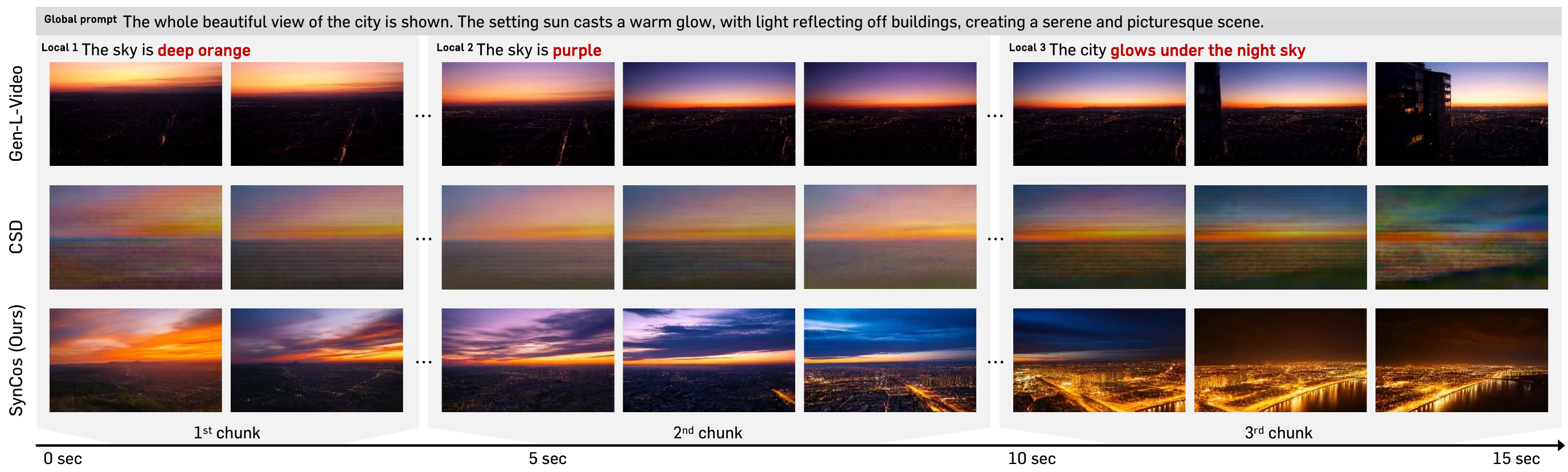}
\vspace{-0.3in}
\captionof{figure}{
\textbf{Qualitative visualization of the ablation study on the three coupled stages of \sname}.
All examples in the second box are 3 times longer in duration compared to the underlying base model.
}\label{fig:abl_stages}
\end{figure*}

%% file: appen/additional_abl.tex
\section{Additional ablations}\label{appen:additional_abl}

\noindent\textbf{Effect of the three coupled stage}. In addition to Section~\ref{sec:motivation} and Figure~\ref{fig:motiv_qual}, we provide further quantitative evaluations (Table~\ref{tab:abl_stages}) and qualitative visualizations (Figure~\ref{fig:abl_stages}) by skipping each stage in \sname to assess the efficacy of its three-stage process in generating high-quality multi-event long videos.

As discussed in Section~\ref{motivation:fusion_genlvideo}, omitting the first stage in \sname, which corresponds to temporal co-denoising with DDIM (\ie, Gen-L-Video), causes denoising paths across chunks to diverge, leading to overlapping artifacts, as shown in Figure~\ref{fig:motiv_qual}.
This results in reduced temporal consistency and frame-wise quality, as quantified in Table~\ref{tab:abl_stages}, due to abrupt changes.
Additionally, this variant often struggles to faithfully follow prompts, as the simple fusion of denoising paths dilutes prompt guidance unique to each chunk.
This issue is evident in the second example of a city transition (Figure~\ref{fig:abl_stages}), where the scene fails to properly reflect changes in glowing, particularly in the third chunk.

Conversely, relying solely on the second stage, corresponding to temporal co-denoising with CSD, degrades image quality significantly.
As shown in Figure~\ref{fig:motiv_qual} and Figure~\ref{fig:abl_stages}, the video exhibits noise-like artifacts, leading to a severe loss of frame-wise quality and prompt fidelity, as quantified in Table~\ref{tab:abl_stages}.
While this approach integrates information across local and distant video frames, it does not produce high-quality long videos, as the resulting video suffers from low image quality, completely failing for high-quality long video generation.

In contrast, \sname effectively couples both stages, leveraging the output of the first stage as a refining source for the second stage, which enhances inter-sample long-range consistency. This integration enables high-quality, long video generation with multiple prompts, achieving smooth transitions, semantic consistency throughout the video, and strong prompt fidelity for each chunk.

\vspace{0.05in}
\noindent\textbf{Ablation study on stride.}
By default, we set the stride (step size between chunks) to 4 for CogVideoX-2B without extensive tuning for each scenario.
For prompts with frequent content changes, reducing the stride improves long-term consistency by increasing information sharing across overlaps.
Users can adjust the stride to balance vibrant changes (larger stride) and stronger synchronization (smaller stride), enhancing content consistency.
Figure~\ref{fig:stride} provides qualitative evidence: while a stride of 4 introduces slight variations in the knight’s appearance, a stride of 1 ensures greater consistency throughout.

\begin{table}[ht]
\centering\small
\includegraphics[width=0.7\linewidth]{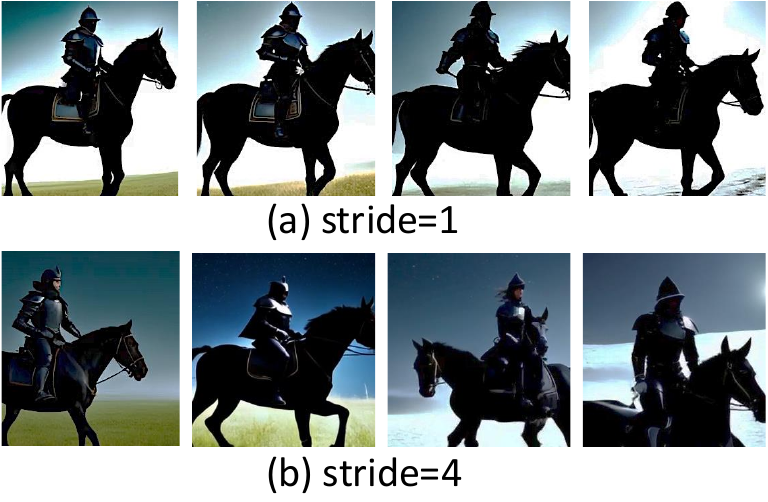}
\vspace{-0.1in}
\captionof{figure}{\textbf{Qualitative ablation study 
on stride.} Reducing the stride enhances content consistency, which is beneficial in scenarios like a knight running as the background transitions from grass to snow.}
\label{fig:stride}
\vspace{-0.15in}
\end{table}

Although reducing the stride increases computation time slightly—from 55 minutes (Table~\ref{tab:comp_time}) to 60 minutes—the impact is minimal.
This is because stride reduction does not affect the second stage optimization time, where most of the computational overhead occurs.
Instead of processing all chunks simultaneously in this stage, \sname uses a minibatch approach, randomly selecting B chunks from the total N chunks at each iteration.
Since B remains the same for both stride 1 and stride 4, the overall optimization cost remains largely unaffected.

%% file: appen/disc_baselines.tex
\input{appen/resources/baselines}

\section{Further discussions on previous work}\label{appen:disc_baselines}

\subsection{Limitations of previous approaches}\label{appen:additional_discussion}

Following the recent success of T2V models, several works have explored extending video diffusion models for longer video generation without additional training. However, we observe that current tuning-free approaches often exhibit undesirable artifacts when applied to recent video diffusion models. In the following sections, we will detail the limitations and failures of these existing methods.

\vspace{0.05in}
\noindent\textbf{Discussion on FIFO-Diffusion.}
In video diffusion models, all frames within a single video chunk are processed at the same timestep during both training and inference.
FIFO-diffusion~\citep{kim2024fifo} proposes a new sampling technique that uses a queue to store a series of consecutive frames, with the timestep increasing for each frame. 
While this approach enables the generation of infinitely long videos, it introduces avoidable discrepancies between the timesteps used during training and those used during inference.
These discrepancies become more pronounced when applied to recent video diffusion models.
This is because as the number of frames processed within a single chunk increases, the timestep gap between frames in the same chunk widens.
For example, FIFO-diffusion is not susceptible to per-frame artifacts on CogVideoX-2B~\cite {yang2024cogvideox}, which encodes 13 frames per chunk.
However, in Open-Sora Plan (v1.3)~\citep{lin2024open}, which encodes 24 frames per chunk, these artifacts become significantly more noticeable, as shown in Figure~\ref{fig:appen_freenoise}. 
We note that in contrast to existing tuning-free long video generation methods, \sname does not introduce training-inference discrepancy and can be seamlessly applied to any video diffusion model.

\vspace{0.05in}
\noindent\textbf{Discussion on FreeNoise.}
The video diffusion models as-is often lack the capability to maintain content consistency across different video chunks beyond a single video chunk.
FreeNoise~\citep{qiu2023freenoise} addresses this limitation by fusing attention features from temporal layers to establish long-range temporal correlations between different chunks.
While this method works effectively with earlier video diffusion models that separate spatial and temporal attention layers (\ie, enabling the fusion of only temporal attention features), we observe that it is not applicable to newer video diffusion models, as illustrated in Figure~\ref{fig:appen_freenoise}.
This limitation arises because recent DiT-based models, widely used in frontier T2V approaches~\citep{yang2024cogvideox, lin2024open}, lack dedicated temporal layers. Instead, these models tokenize an entire video chunk into patches and apply attention across all patches. 
As a result, fusing all attention features without additional considerations severely degrades performance, demonstrating that naive fusion techniques are unsuitable for DiT-based models in long video generation.

\subsection{\texorpdfstring{Additional comparisons with \\architecture-specific approaches}{Additional comparisons with architecture-specific approaches}}\label{appen:additional_comp}

In our main experiment (Section~\ref{sec:exp}), we primarily compare \sname with architecture-compatible, tuning-free methods for multi-event long video generation.
These approaches remain compatible with newer diffusion models, leveraging backbone improvements for enhanced generation quality.
However, several existing methods~\citep{lu2025freelong, tan2024video, cai2024ditctrl} are restricted to specific diffusion backbones, making them incompatible with newer or alternative models and limiting their performance to existing architectures.

Nonetheless, to ensure a comprehensive and fair comparison, we implement \sname using the respective architectures of existing tuning-free methods and evaluate its performance in Table~\ref{tab:comp2}.
We compare \sname with Video-Infinity~\citep{tan2024video} by applying \sname to VideoCrafter2~\citep{wang2023videofactory}, a U-Net-based video diffusion model. 
Additionally, to compare against DitCtrl~\citep{cai2024ditctrl} (built on Multi-Modal Diffusion Transformer (MM-DiT)), we evaluate \sname under the same backbone of CogVideoX-2B~\citep{yang2024cogvideox}.

Notably, \sname significantly outperforms all baselines in temporal consistency and video quality while achieving comparable prompt fidelity, demonstrating its robustness across various diffusion backbones.

\begin{table}[h]
\centering\small
\caption{\textbf{Quantitative comparison with architecture-specific baselines}. *Abbreviations: subject consistency (SC), background consistency (BC), aesthetic quality (AQ), and prompt fidelity (PF).
}
\vspace{-0.1in}
\label{tab:comp2}
\resizebox{0.8\linewidth}{!}{
\begin{tabular}{c l c c c c}
\toprule
 & & \multicolumn{2}{c}{Temporal}  & \multicolumn{1}{c}{Frame} & \multicolumn{1}{c}{Semantics} \\
    \cmidrule(lr){3-4} \cmidrule(lr){5-5}  \cmidrule(lr){6-6} 
    Backbone & Method 
    & SC $\uparrow$ 
    & BC $\uparrow$
    & AQ $\uparrow$
    & PF $\uparrow$ \\
    \midrule

    \multirow{2.5}{*}{VideoCrafter2~\citep{wang2023videofactory}} 
    & Video-Infinity~\citep{tan2024video} & 0.879 & 0.943 & 0.645 & \textbf{0.365} \\
    \cmidrule(lr){2-6} 
    & \textbf{SynCoS} & \textbf{0.911} & \textbf{0.947} & \textbf{0.648} & \textbf{0.365} \\

    \midrule

    \multirow{2.5}{*}{CogVideoX-2B~\citep{yang2024cogvideox}}     
    & DitCtrl~\citep{cai2024ditctrl} &  0.821 & 0.916  & 0.635 & \textbf{0.394} \\
    \cmidrule(lr){2-6} 
    & \textbf{SynCoS} & \textbf{0.864} & \textbf{0.927} & \textbf{0.643} & 0.381 \\

    \bottomrule
\end{tabular}
}
\vspace{-0.1in}
\end{table}

\vspace{0.05in}
\noindent\textbf{Comparisons with autoregressive generation using I2V.}
Long videos can also be generated using image-to-video (I2V) models by generating a single video chunk with T2V and then conditioning the last frame of the previous chunk to the I2V model.
For two reasons, we do not directly include autoregressive generation using the I2V model in the main paper.
First, using an I2V model requires a different prompt structure. Unlike our structured prompts, which focus on describing changes in local chunks, I2V models require explicit descriptions of transitions between chunks. 
For example, to generate an artistic video of a butterfly on a flower across changing seasons, our structured prompts might include: (1) ``In spring, a butterfly is on a flower,'' and (2) ``In summer, a butterfly is on a flower.'' In contrast, I2V prompts would need to explicitly describe transitions, such as: (1) ``In spring, a butterfly is on a flower,'' and (2) ``The season changes from spring to summer, and the butterfly is on a flower.'' 
Second, the base models used in our experiments do not support I2V generation.

Nonetheless, to comprehensively analyze long video generation, we compare \sname with an I2V model (CogVideo-5X-I2V) that uses a backbone supporting I2V generation and appropriately tuned prompts.
As shown in Figure~\ref{fig:appen_i2v}, I2V-based autoregressive generation enables smooth scene transitions (e.g., a Chihuahua floating in space transitioning to the Chihuahua swimming in the ocean). However, it often struggles to introduce new objects (e.g., a brown squirrel and a white squirrel, or a Chihuahua with fish), limiting its ability to generate long videos that naturally add or change components over time.

%% file: appen/resources/baselines.tex
\begin{figure*}[t]
\centering\small

\begin{subfigure}[t]{\textwidth}
    \centering
    \includegraphics[width=\textwidth]{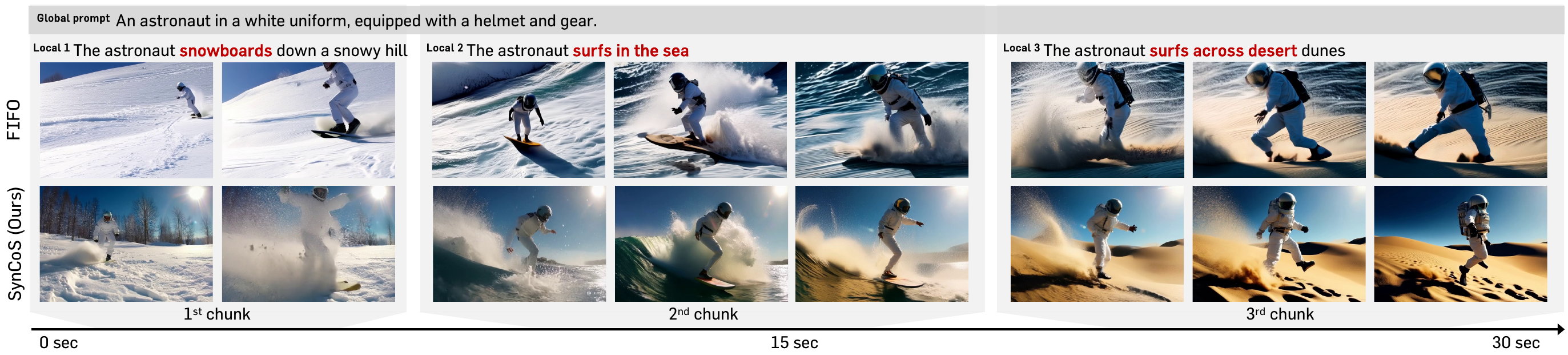}
    \vspace{-0.2in}
    \caption*{(a) Long video generation on CogVideoX-2B, where a single video chunk consists of 26 frames.}
\end{subfigure}

\vspace{0.1in}

\begin{subfigure}[t]{\textwidth}
    \centering
    \includegraphics[width=\textwidth]
    {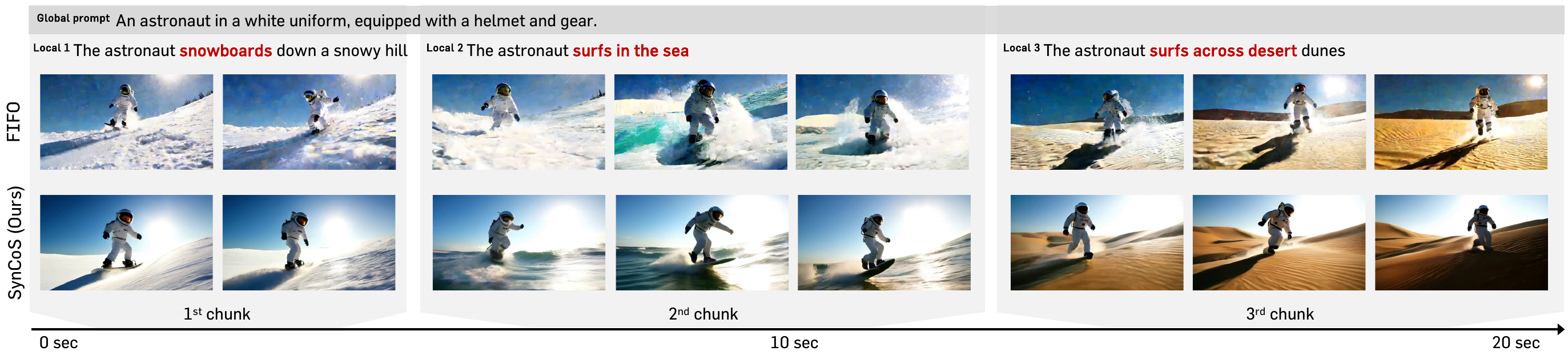}
    \vspace{-0.2in}
    \caption*{(b) Long video generation on Open-Sora Plan (v1.3), where a single video chunk consists of 49 frames.}
\end{subfigure}
\vspace{-0.1in}
\caption{\textbf{Long video generation results}. All generated videos are 4 times longer than the underlying base model.
FIFO significantly suffers from noise-like artifacts on Open-Sora Plan (v1.3) due to inevitable training-inference discrepancy in their design.
}\label{fig:appen_fifo}

\vspace{0.3in}

\includegraphics[width=\textwidth]{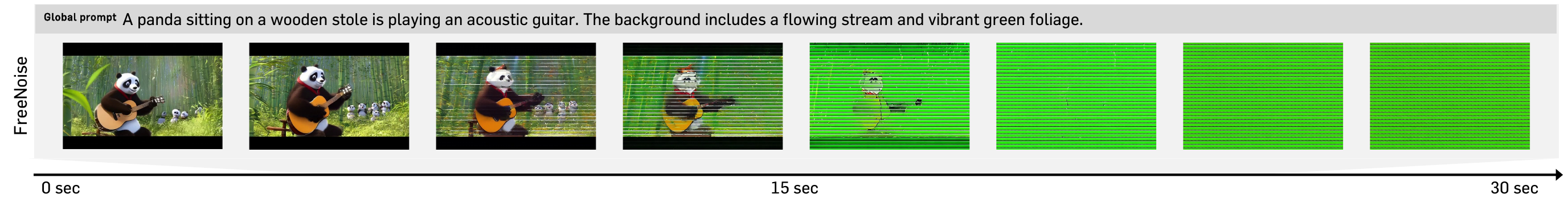}
\vspace{-0.25in}
\caption{\textbf{Long video generation result of FreeNoise on CogVideoX-2B}. This generated video is 4 times longer than the underlying base model. Each frame is generated well in the early frames, where no fusion is applied. However, as soon as the fusion of attentional features is applied, the generated video shows stagnant results of repeated object motion without any scene changes, eventually leading to the entire failure of video generation.}\label{fig:appen_freenoise}

\vspace{0.3in}

\includegraphics[width=\textwidth]{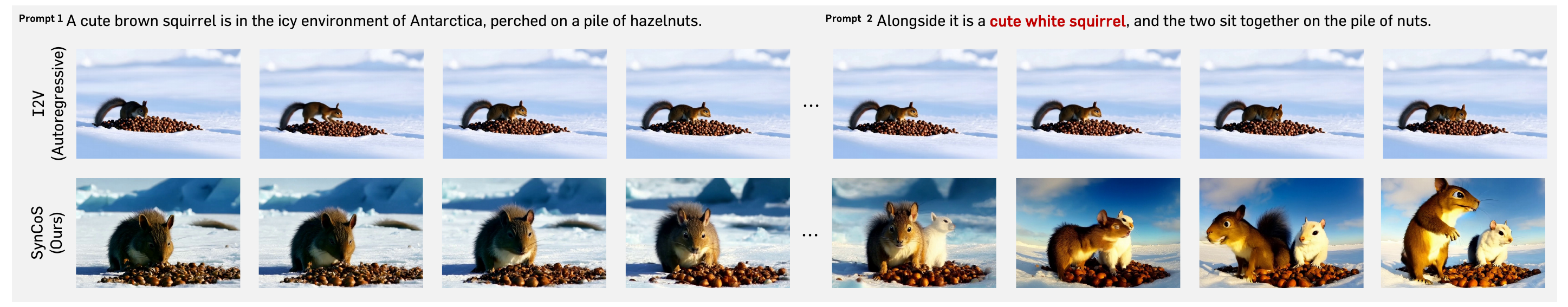}
\includegraphics[width=\textwidth]{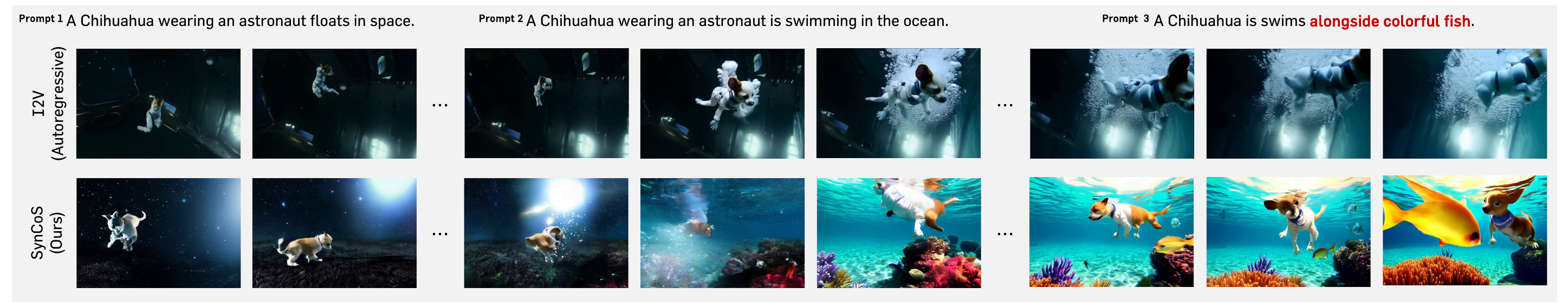}
\vspace{-0.25in}
\caption{\textbf{Qualitative comparison} with autoregressive generation using image-to-video (I2V) model for long videos.
While autoregressive generation with I2V models effectively handles scene transitions, it often struggles to introduce new components into the video.}
\label{fig:appen_i2v}
\end{figure*}

%% file: appen/pseudo_code.tex
\clearpage
\section{Pseudo-code implementation of each stage.}\label{appen:pseudo}
We describe each stage with pseudo-code implementations.
In the first stage of temporal co-denoising with DDIM, the video is divided into overlapping chunks, where each chunk undergoes a diffusion forward pass and a DDIM update to compute $\mathbf{x}_0^{(i)}$.
The chunks are then fused to obtain $\mathbf{x}'_0$, as detailed in Figure~\ref{fig:appen_code_first}.
In the second stage, the fused $\mathbf{x}'_0$ is further refined by re-dividing it into overlapping chunks and applying CSD-based optimization.
While we use CSD-based optimization to enforce global coherence, it differs from the original CSD, as the second stage is adjusted using a grounded timestep and a fixed baseline noise, ensuring synchronous coupling across all three stages.
The iterative refinement process is illustrated in Figure~\ref{fig:appen_code_second}.
Lastly, the refined $\mathbf{x}'_0$ from the second stage is converted using baseline noise to prepare for subsequent steps in the diffusion pipeline.

\begin{figure*}[htb]
\centering\small
\resizebox{0.65\textwidth}{!}{%
\begin{lstlisting}[language=Python, basicstyle=\ttfamily\scriptsize, numbers=left, breaklines]
def first_stage(
    videos,                      # video tensor of shape [T, C, H, W]
    prompt_embeds_for_chunks,      # Text prompt embeddings for each chunk
    model,                        # The denoising model
    scheduler,                    # Scheduler to refine videos
    guidance_scale,               # Scale factor for classifier-free guidance
    timestep,                     # Current timestep in the diffusion process
    stride                        # Stride size for dividing videos
):
    
    # Initialize list to store video chunks and count tensor
    videos_for_chunks = []
    count = torch.zeros_like(videos)  

    # Split videos into strides for diffusion forward
    for start in range(0, videos.shape[0], stride):
        # Determine the end of the current stride
        end = start + stride

        # Slice the videos for the current stride
        chunks = videos[start:end]

        # Increment count for the covered range
        count[start:end] += 1

        # Append the current stride to the list for processing
        videos_for_chunks.append(chunks)

    # Concatenate all processed video chunks into a single tensor for inference
    videos_for_chunks_input = torch.cat(videos_for_chunks, dim=0)

    # Diffusion forward
    noise_pred = predict_noise(model, videos_for_chunks_input, prompt_embeds_for_chunks, timestep)

    # Apply classifier-free guidance to refine noise predictions
    noise_pred = apply_classifier_free_guidance(noise_pred, guidance_scale)

    # Perform a scheduler step to update videos and extract the denoised sample
    res = scheduler_step(scheduler, noise_pred, timestep, videos_for_chunks)
    videos_x0 = res.pred_original_sample

    # Initialize the aggregated result tensor
    value_x0 = torch.zeros_like(videos)

    # Aggregate the results into `value_x0`
    idx = 0
    for start in range(0, videos.shape[0], stride):
        # Define the range for the current stride
        start + stride

        # Accumulate the processed values for the corresponding stride
        value_x0[start:end] += videos_x0[idx]
        idx += 1

    # Compute the final denoised prediction by fusing
    # Wherever `count > 0`, compute the averaged value, else fallback to `value_x0`
    pred_original_sample = torch.where(count > 0, value_x0 / count, value_x0)

    return pred_original_sample
\end{lstlisting}
}
\caption{\textbf{Pseudo-code implementation of the first stage of \sname}.}
\label{fig:appen_code_first}
\end{figure*}

\begin{figure*}[htb]
\centering\small
\resizebox{0.65\textwidth}{!}{%
\begin{lstlisting}[language=Python, basicstyle=\ttfamily\scriptsize, numbers=left, breaklines]
def second_stage(
        pred_original_sample,  # Initial video prediction
        prompt_embeds_for_chunks,      # Text prompt embeddings for each chunk
        timestep,              # A grounded timestep
        cfg_scale,             # Guidance scale for classifier-free guidance
        opt                    # Optimization hyperparameters and options
    ):

    # Initialize target predictions with gradients enabled
    tgt_pred_sample = pred_original_sample.clone().detach()
    tgt_pred_sample.requires_grad = True

    # Optimization hyperparameters
    lr = opt.lr
    wd = opt.wd
    decay_iter = opt.decay_iter
    decay_rate = opt.decay_rate
    num_steps = opt.num_steps
    batch_size = opt.batch_size
    stride = opt.stride

    # Initialize optimizer and learning rate scheduler
    optimizer = opt.optimizer([tgt_pred_sample], lr=lr, weight_decay=wd)
    scheduler = opt.scheduler(optimizer, step_size=decay_iter, gamma=decay_rate)

    # Fix baseline noise for the entire tensor
    base_noise_all = torch.randn_like(tgt_pred_sample)

    for step in range(num_steps):
        optimizer.zero_grad()

        # Initialize tensors for counting and score accumulation
        count = torch.zeros_like(tgt_pred_sample, dtype=tgt_pred_sample.dtype)
        score_value = torch.zeros_like(tgt_pred_sample, dtype=tgt_pred_sample.dtype)

        # Initialize lists to store videos and noise chunks
        videos_for_chunks = []
        basenoise_chunks = []
        prompt_embeds_chunks = []

        # Determine valid subset indices for this iteration
        num_chunks = (tgt_pred_sample.shape[0] + stride - 1) // stride
        subset_indices = torch.randperm(num_chunks)[:batch_size]

        # Process selected chunks
        for i, start in enumerate(range(0, tgt_pred_sample.shape[0], stride)):
            if i not in subset_indices:
                continue

            # Determine the end of the current stride
            end = start + stride

            # Slice the videos for the current stride and add noise
            chunks = tgt_pred_sample[start:end]
            noisy_chunks = add_noise(chunks, base_noise_all[start:end], timestep)

            basenoise_chunks.append(base_noise_all[start:end].unsqueeze(0))
            prompt_embeds_chunks.append(prompt_embeds_for_chunks[i])

            # Update count and append to chunks list
            count[start:end] += 1
            videos_for_chunks.append(noisy_chunks.unsqueeze(0))

        # Concatenate videos for model inference
        videos_for_chunks_input = torch.cat(videos_for_chunks, dim=0)
        basenoise_chunks = torch.cat(basenoise_chunks, dim=0)
        prompt_embeds_chunks = torch.cat(prompt_embeds_chunks, dim=0)
        
        # Predict noise using the model
        noise_pred = predict_noise(opt.model, videos_for_chunks_input, prompt_embeds_chunks, timestep)

        # Apply classifier-free guidance
        noise_pred = apply_classifier_free_guidance(noise_pred, cfg_scale)

        # Compute CSD scores
        scores = calculate_csd_loss(videos_for_chunks_input, noise_pred, basenoise_chunks)

        # Aggregate scores back to the appropriate positions
        for i, start in enumerate(range(0, tgt_pred_sample.shape[0], stride)):
            if i not in subset_indices:
                continue

            # Determine the end of the current stride
            end = start + stride

            # Accumulate scores for the current range
            score_value[start:end] += scores[start:end]

        # Compute gradients using scores normalized by count
        grad_all = torch.where(count > 0, score_value / count, score_value)
        tgt_pred_sample.backward(gradient=grad_all)

        # Perform optimization step
        optimizer.step()
        scheduler.step()

    # Detach and return the updated target prediction
    tgt_pred_sample = tgt_pred_sample.clone().detach().requires_grad_(False)
    
    return tgt_pred_sample, base_noise_all
\end{lstlisting}
}
\caption{\textbf{Pseudo-code implementation of the second stage of \sname}.}
\label{fig:appen_code_second}
\end{figure*}

%% file: appen/qual_comp_additional.tex
\section{Additional qualitative results}\label{appen:qual}
\begin{figure*}[t]
    \centering


    \begin{subfigure}[t]{\textwidth}
    \includegraphics[width=\textwidth]{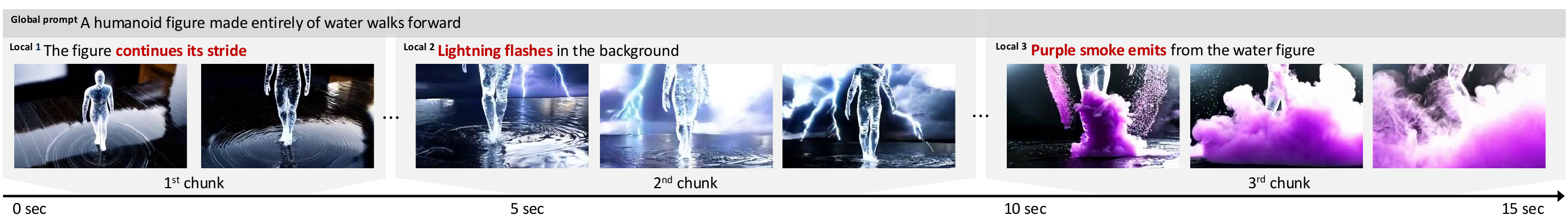}
    \includegraphics[width=\textwidth]{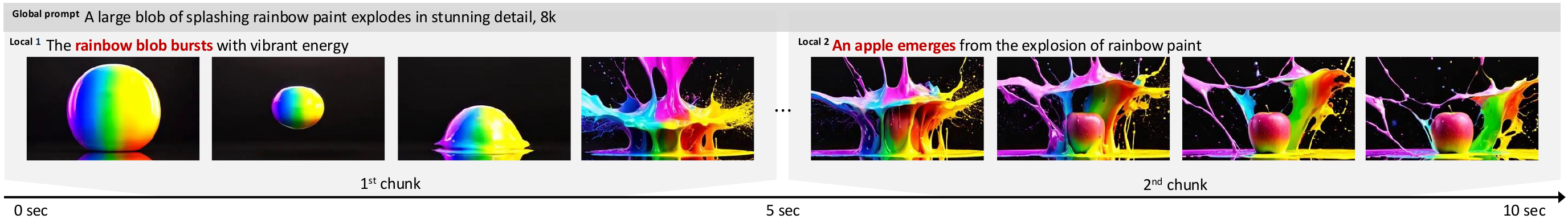}
    \includegraphics[width=\textwidth]{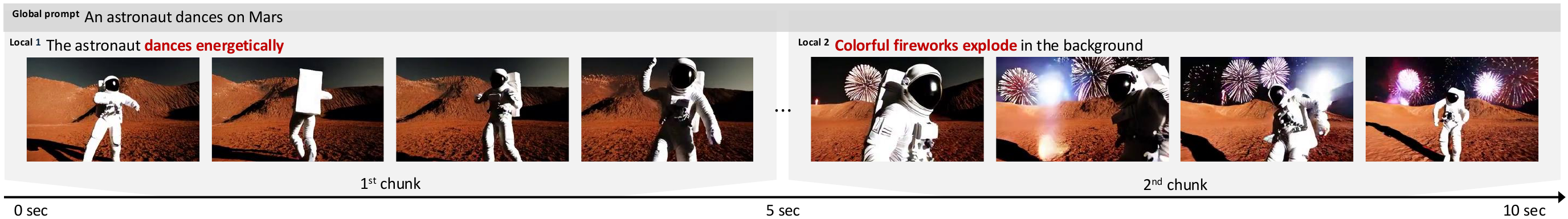}
    \caption*{(a) Cinematic effects}
    \end{subfigure}

    \vspace{0.4in}
    
    \begin{subfigure}[t]{\textwidth}
    \includegraphics[width=\textwidth]{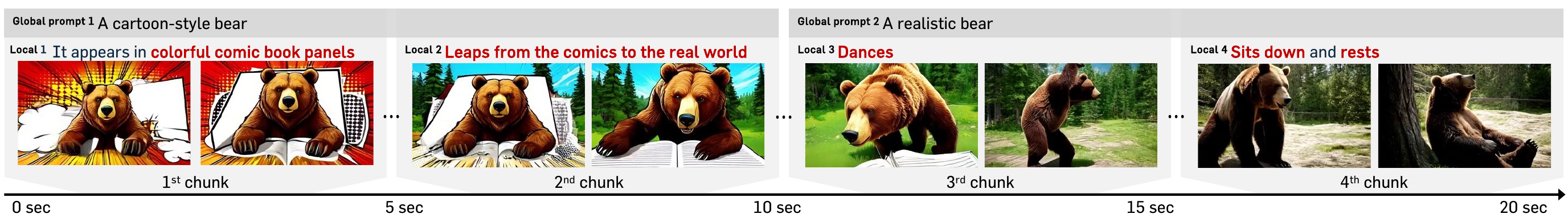}
    \caption*{(b) Storytelling}
    \end{subfigure}

    \vspace{0.4in}

    \begin{subfigure}[t]{\textwidth}
    \includegraphics[width=\textwidth]{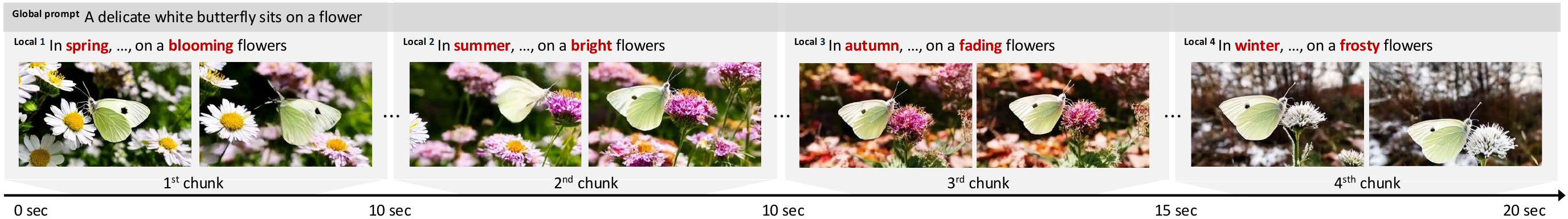}
    \includegraphics[width=\textwidth]{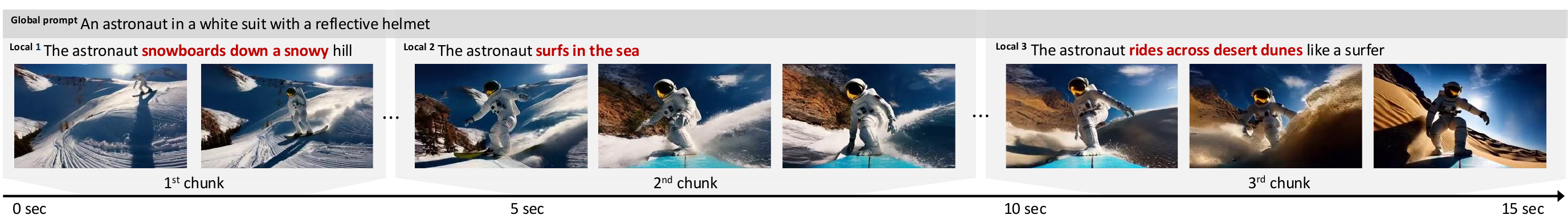}
    \caption*{(c) Background changes}
    \end{subfigure}

    \caption{\textbf{Multi-event long video generation results} showcasing challenging scenarios, including cinematic effects, storytelling, and background changes.
    Each example is 2-4 times longer in duration compared to the underlying base model, resulting in 11 to 21-second videos at 24 fps, with a total of 256 to 512 frames.}\label{fig:add_qual_1}

\end{figure*}

\begin{figure*}[t]
    \centering

    \begin{subfigure}[t]{\textwidth}
    \includegraphics[width=\textwidth]{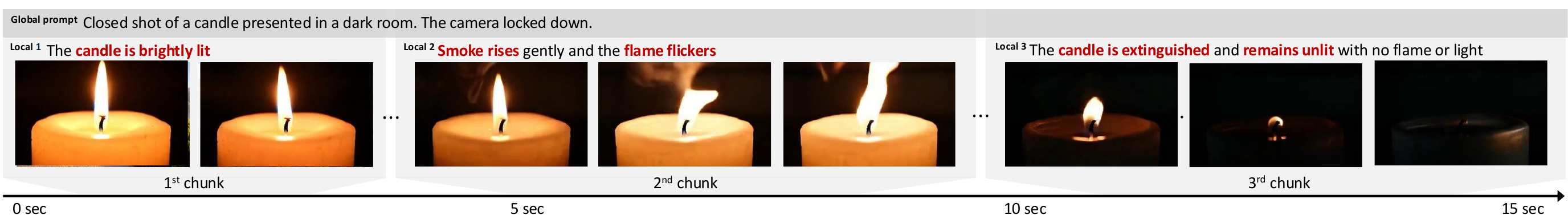}
    \caption*{(d) Physical transformation}
    \end{subfigure}
    
    \vspace{0.4in}
    
    \begin{subfigure}[t]{\textwidth}
    \includegraphics[width=\textwidth]{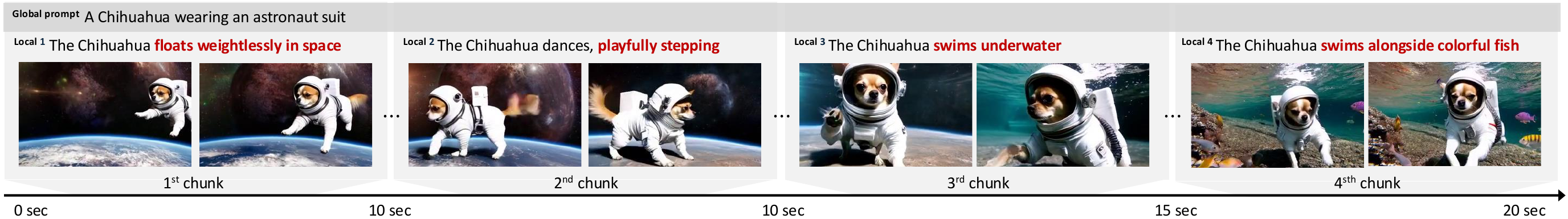}
    \includegraphics[width=\textwidth]{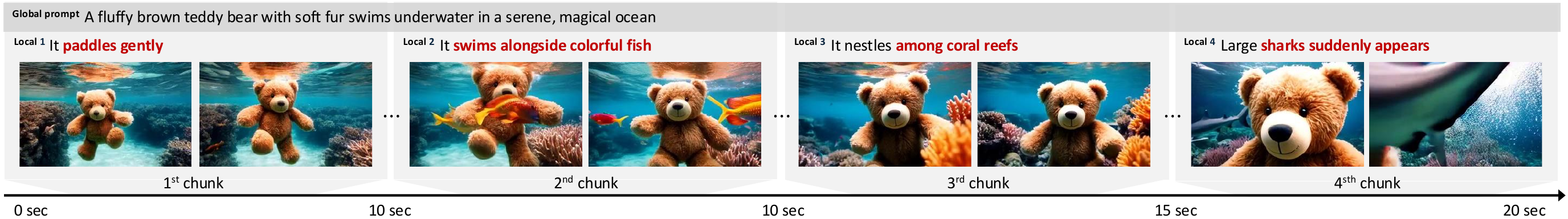}
    \caption*{(e) Complex scene transitions}
    \end{subfigure}

    \vspace{0.4in}

    \begin{subfigure}[t]{\textwidth}
    \includegraphics[width=\textwidth]{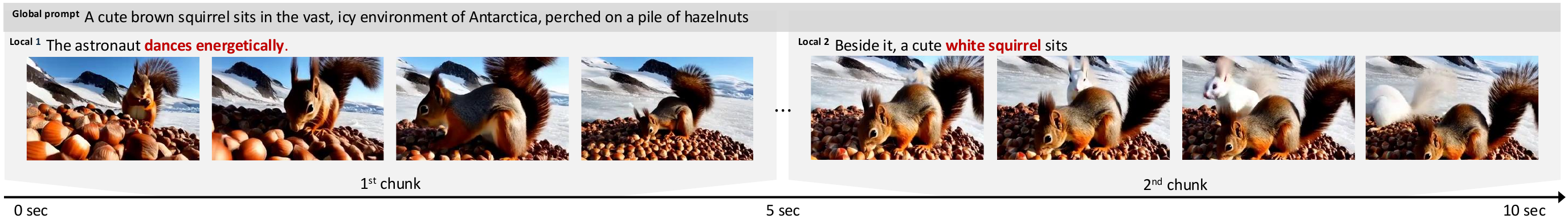}
    \includegraphics[width=\textwidth]{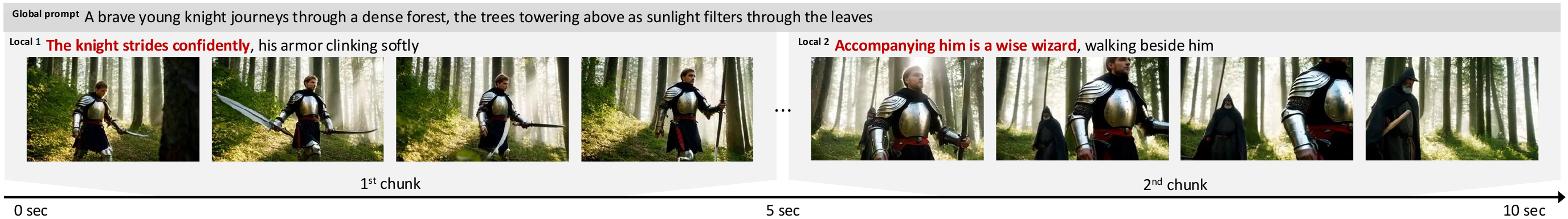}
    \caption*{(f) Compositional generation}
    \end{subfigure}

    \caption{\textbf{Multi-event long video generation results} showcasing challenging scenarios, including physical transformation, complex scene transitions, and compositional generation.
    Each example is 2-4 times longer in duration compared to the underlying base model, resulting in 11 to 21-second videos at 24 fps, with a total of 256 to 512 frames.}\label{fig:add_qual_2}
    
\end{figure*}

We consider long videos of multiple events with the following challenging scenarios: object motion control, cinematic effects, storytelling, camera control, background changes, physical transformations, complex scene transitions, and compositional generation. 
For generated videos handling object motion control and camera control scenes, please refer to Figure~\ref{fig:teaser}, and other video examples are visualized in Figure~\ref{fig:add_qual_1} and Figure~\ref{fig:add_qual_2}.
To view all generated videos, refer to our project page: \url{https://syncos2025.github.io/}.


